\documentclass[10pt,journal,compsoc]{IEEEtran}

\usepackage{xcolor,soul,framed} %,caption

\colorlet{shadecolor}{yellow}
\usepackage[pdftex]{graphicx}
\graphicspath{{../pdf/}{../jpeg/}}
\DeclareGraphicsExtensions{.pdf,.jpeg,.png}

\usepackage[cmex10]{amsmath}
\usepackage{array}
\usepackage{mdwmath}
\usepackage{mdwtab}
\usepackage{eqparbox}
\usepackage{url}
\usepackage{color} 
\usepackage{cite}
\usepackage{amssymb,amsfonts}
\usepackage{multirow}
\usepackage{subcaption} 
\usepackage{float} 
\usepackage{algorithm}
\usepackage{algpseudocode}
\usepackage{color}
\usepackage{booktabs}
\usepackage{lineno}
\usepackage{hyperref}
\usepackage{longtable}

\usepackage{enumitem}
\begin{document}
%\bstctlcite{IEEEexample:BSTcontrol}
\title{Spatio-Temporal Graph Neural Networks for Predictive Learning in Urban Computing: \\A Survey}
\author{Guangyin Jin, Yuxuan Liang,~\IEEEmembership{Member,~IEEE}, Yuchen Fang, Zezhi Shao, \\ Jincai Huang, Junbo Zhang,~\IEEEmembership{Senior Member,~IEEE}, Yu Zheng,~\IEEEmembership{Fellow,~IEEE}
%\IEEEcompsocitemizethanks{\IEEEcompsocthanksitem F.~Li,  M.~Wang are with Beijing National Research Center for Information Science and Technology (BNRist), Department of Electronic Engineering, Tsinghua University, Beijing 100084, China.\protect\\
%E-mail: lifx19@mails.tsinghua.edu.cn, wmd20@mails.tsinghua.edu.cn
%\protect\\
\IEEEcompsocitemizethanks{\IEEEcompsocthanksitem G.Y Jin is with National Innovative Institute of Defense Technology, Beijing, China, and National University of Defense Technology, Changsha, China. E-mail: jinguangyin96@foxmail.com, 
\IEEEcompsocthanksitem Y.X~Liang is with Intelligent Transportation Thrust, Hong Kong University of Science and Technology (Guangzhou), Guangzhou, China. Email: yuxliang@outlook.com
\IEEEcompsocthanksitem Y.C~Fang is with the School of Computer Science and Engineering, University of Electronic Science and Technology of China, Chengdu, China.
E-mail: fyclmiss@gmail.com
\IEEEcompsocthanksitem Z.Z~Shao is with  Institute
of Computing Technology, University of Chinese Academy of Sciences, Beijing, China. E-mail: shaozezhi19b@ict.ac.cn
\IEEEcompsocthanksitem J.C~Huang is with College of Systems Engineering, National University of Defense Technology, Changsha, China. E-mail: huangjincai@nudt.edu.cn
\IEEEcompsocthanksitem J.B~Zhang and Y.~Zheng are with JD Intelligent Cities Research and JD iCity, JD Technology, Beijing, China. E-mail: msjunbozhang@outlook.com, msyuzheng@outlook.com
\IEEEcompsocthanksitem Y.X~Liang is the corresponding author.
}

}

% The paper headers
% \markboth{IEEE TRANSACTIONS ON MICROWAVE THEORY AND TECHNIQUES, VOL.~60, NO.~12, DECEMBER~2012
% }{Roberg \MakeLowercase{\textit{et al.}}: High-Efficiency Diode and Transistor Rectifiers}

% ====================================================================

\markboth{}
{Shell \MakeLowercase{\textit{et al.}}: Bare Demo of IEEEtran.cls for Computer Society Journals}

\IEEEtitleabstractindextext{%
\begin{abstract}
With recent advances in sensing technologies, a myriad of spatio-temporal data has been generated and recorded in smart cities. Forecasting the evolution patterns of spatio-temporal data is an important yet demanding aspect of urban computing, which can enhance intelligent management decisions in various fields, including transportation, environment, climate, public safety, healthcare, and others. Traditional statistical and deep learning methods struggle to capture complex correlations in urban spatio-temporal data. To this end, Spatio-Temporal Graph Neural Networks (STGNN) have been proposed, achieving great promise in recent years. STGNNs enable the extraction of complex spatio-temporal dependencies by integrating graph neural networks (GNNs) and various temporal learning methods. 
In this manuscript, we provide a comprehensive survey on recent progress on STGNN technologies for predictive learning in urban computing. Firstly, we provide a brief introduction to the construction methods of spatio-temporal graph data and the prevalent deep-learning architectures used in STGNNs. We then sort out the primary application domains and specific predictive learning tasks based on existing literature. Afterward, we scrutinize the design of STGNNs and their combination with some advanced technologies in recent years. Finally, we conclude the limitations of existing research and suggest potential directions for future work.

\end{abstract}

}

\maketitle

\IEEEdisplaynontitleabstractindextext

\IEEEpeerreviewmaketitle

% For peer review papers, you can put extra information on the cover
% page as needed:
% \ifCLASSOPTIONpeerreview
% \begin{center} \bfseries EDICS Category: 3-BBND \end{center}
% \fi
%
% For peerreview papers, this IEEEtran command inserts a page break and
% creates the second title. It will be ignored for other modes.
%\IEEEpeerreviewmaketitle

% ====================================================================
% ====================================================================
% ====================================================================

% === I. INTRODUCTION =============================================================
% =================================================================================
\section{Introduction}\label{sec:intro}
With the rapid advancement of sensing and data stream processing technologies, vast amounts of data in urban systems have been efficiently collected and stored. This has laid the foundation for the era of urban computing, which aims to understand the urban patterns and dynamics from different application domains where the big data explodes, such as transportation, environment, climate, etc. 
Predictive learning is a typical supervised learning paradigm that learns from historical data to forecast future trends.
According to urban computing theories~\cite{zheng2014urban}, predictive learning based on massive urban data is the most important loop, forming the foundation for intelligent decision-making, scheduling, and management in smart cities. In addition, the predictability of urban big data can also provide the possibility for the development of some new technologies such as digital twin cities and metaverse~\cite{wang2016acp}.

% \begin{figure}[!t]
% \centering
% \includegraphics[width=0.48 \textwidth]{f25.png}
% \caption{An example of spatio-temporal heterogeneity. (a) Districts with different functionality in an urban network. (b) Statistic of crowd flow data in different nodes from various districts. Despite all nodes exhibiting evident peak patterns, there are notable variations in crowd flow values across nodes located in different districts. Conversely, nodes situated within the same district display similar values even if they differ in their location, \emph{e.g.}, the case with node 3\&4.}
% \label{fig:example} % FIG
% \vspace{-1em}
% \end{figure}

The majority of urban data is spatio-temporal, representing that it not only pertains to spatial locations but also changes over time. Within urban systems, spatio-temporal data exhibits ubiquitous properties of \emph{correlation} and \emph{heterogeneity}~\cite{wang2020deep}. Correlation refers to the data being auto-correlated not only in the temporal dimension, but also in the spatial dimension. Heterogeneity is a property of spatio-temporal data wherein it displays varying patterns across different temporal or spatial ranges.%, as illustrated in Figure~\ref{fig:example}. 
The complex nature of the above characteristics has resulted in an increased difficulty in feature engineering. As a consequence, some methods that performed well in traditional time series forecasting, such as Support Vector Regression (SVR)~\cite{drucker1996support}, Random Forest (RF)~\cite{breiman2001random}, and Gradient Boosting Decision Tree (GBDT)~\cite{natekin2013gradient}, are less effective in achieving accurate prediction results. In the past decade,  the rapid development of deep learning technologies has led to the emergence of hybrid neural networks based on Convolutional Neural Networks (CNN)~\cite{gu2018recent} and Recurrent Neural Networks (RNN)~\cite{yu2019review}. These hybrid networks (\emph{e.g.}, ConvLSTM~\cite{shi2015convolutional}, PredRNN~\cite{wang2017predrnn}) have been increasingly applied to predictive learning of urban spatio-temporal data and have shown significant advantages. However, the major limitation of these methods is their inability to learn directly from non-Euclidean data existing in urban systems, such as vehicle flows over road networks, traffic on route networks, and entities in urban knowledge graphs. 

\begin{figure}[!t]
\vspace{-3mm}
\centering
\includegraphics[width=0.42 \textwidth]{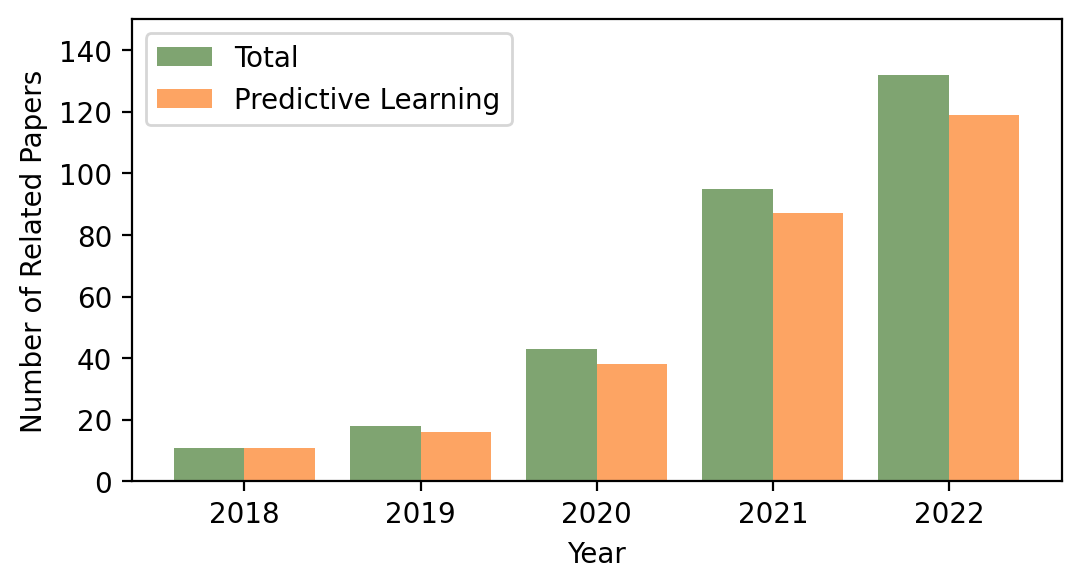}
\caption{The publication trend of STGNN-related papers in Google Scholar over the past five years. The blue bars represent the total number of relevant publications and the red bars denote those focusing on predictive learning tasks.}
\label{fig:intro} % FIG
\vspace{-2mm}
\end{figure}

Over the past few years, there have been significant breakthroughs in representation learning of non-Euclidean data through deep learning techniques, particularly Graph Neural Networks (GNN)~\cite{kipf2016semi}. This has paved the way for predictive learning of diverse and intricate urban data. Given the spatio-temporal characteristics of urban data, such as traffic flows, a line of studies integrated GNNs with various temporal learning methods to capture dynamics in both space and time dimension~\cite{wang2020deep}. This type of hybrid neural architecture is generally known as \textbf{Spatio-Temporal Graph Neural Network (STGNN)}. Recently, STGNNs have been widely used for predictive learning scenarios in urban computing, including transportation, environment, public safety, health, energy, economy, and other fields. Using the search engine of Google Scholar, we perform meticulous keyword searches and tally the relevant paper publications in the past five years. As depicted in Figure~\ref{fig:intro}, we can witness a notable surge in the number of relevant papers on STGNN year by year. In 2018, there are fewer than 20 papers, while in 2022, the number reaches nearly 140. This trend of progress indicates that STGNN-related applications have emerged as a highly sought-after research area in recent years. It is worth noting that a majority of these publications concentrated on predictive learning tasks.

\textbf{Related Surveys.} In recent years, there have been a few related surveys on the applications of STGNN-based predictive learning techniques across different fields. Wang et al.~\cite{wang2020deep} conducted a review of deep learning techniques for spatio-temporal data mining involving a series of STGNNs in predictive learning up to 2020. There were also several surveys~\cite{ye2020build,bui2022spatial,jiang2022graph} investigating the blossom of STGNNs in transportation domains. To be specific, ~\cite{ye2020build} analyzed multiple practical problems and revisited related works about prediction, detection, and control problems in urban traffic systems. Bui et al.~\cite{bui2022spatial} and Jiang et al.~\cite{jiang2022graph} focused on the latest STGNN technologies in traffic forecasting tasks. 
% Gao et al.~\cite{gao2022generative} summarized a wide range of applications using Generative Adversarial Networks (GAN) for spatio-temporal data learning, including some approaches combined with spatio-temporal graph data. 
In recent months, there has been a small number of works~\cite{sahili2023spatio,li2023graph} surveying applications of STGNNs, but their topics are broad and do not focus on predictive learning in urban computing.

\textbf{Our Contributions.} In contrast to prior surveys, the contributions of our survey lie in four aspects:
\begin{itemize}[leftmargin=*]

\item To our knowledge, this is the first comprehensive survey to systematically review recent studies that use STGNNs for predictive learning in urban computing. We scrutinize the progress of STGNN from both application and methodology perspectives based on extensive literature. 

\item We categorize the primary application domains as well as particular predictive learning tasks of STGNNs in urban computing, and sort out a list of public datasets attached with the previous works on STGNNs.

\item We provide an in-depth analysis of existing STGNN methods for temporal learning, spatial learning, and spatio-temporal fusion. We further examine some recently emerging approaches that integrated STGNN with other advanced learning frameworks. 

\item We summarize the challenges shared by STGNNs for predictive learning tasks in urban computing and suggest future directions for addressing these challenging issues.
\end{itemize}

\textbf{Organization.} The rest is organized as follows. Section~\ref{sec:stgc} illustrates how to construct spatio-temporal graphs based on prior knowledge. In Section~\ref{sec:tax}, a taxonomy of STGNNs for predictive learning in urban computing is presented. Section~\ref{sec:domain} overviews various predictive learning tasks from different domains that can be addressed by STGNNs. Section~\ref{sec:basic_arch} delineates the fundamental deep learning architectures commonly used in STGNNs. Section~\ref{sec:improved} and Section~\ref{sec:advanced} delve into an in-depth analysis of the neural architecture design methods of STGNNs and popular advanced techniques that can be combined, respectively. Section~\ref{sec:challenge} further highlights the limitations of existing works and suggests future directions. Finally, we conclude this survey in Section~\ref{sec:conclusion}.
To facilitate a quick understanding of the formulas in the paper, we have also compiled a list of symbols that encompasses the most commonly used symbol notations, as shown in Table~\ref{tab:notation}.

% \section{Preliminary}
% In this section, we sort out some background knowledge for STGNN, including spatio-temporal graph construction, graph neural networks, recurrent neural networks, temporal convolutional neural networks and self-attention neural networks. 

\begin{table}[htbp]
	\caption{Summary of symbol notations.}
 \vspace{-3mm}
	\footnotesize
	\begin{center}
		\begin{tabular}{|c|p{0.75 \linewidth}|}			
			\hline
			\multicolumn{1}{|c|}{\textbf{Notation}} & \multicolumn{1}{|c|}{\textbf{Definition}} \\ 
			\hline
                $\mathcal{G}_{t}$  & Input ST-Graph at time $t$  \\  \hline
                $A_{t}$ & Adjacency matrix of input ST-Graph at time $t$\\ \hline
                $\mathcal{E}_{t}$ & Edge set of input ST-Graph at time $t$ \\ \hline              
                $a_{ij}^{t}$ & Connection weight between nodes i and j at time $t$ \\ \hline
                ${\mathcal X}_{t}$ &  Input node features at time $t$ \\ \hline
                $x_{i}^{t}$ &  Input features of node i at time $t$ \\ \hline                
                $\boldsymbol{f}(\cdot)$ & An arbitrary nonlinear function \\ \hline
                $H_t$ & The hidden state of input node features at time $t$\\ \hline
                $\star$ & Convolution operator \\ \hline	
                $\odot$ & Element-wise product operator \\ \hline
		\end{tabular}
		\label{tab:notation}
	\end{center}
 \vspace{-5mm}
\end{table}

\section{Spatio-Temporal Graph Construction}~\label{sec:stgc}
\begin{figure}[!t] 
\centering
\vspace{-3mm}
\includegraphics[width=0.45 \textwidth]{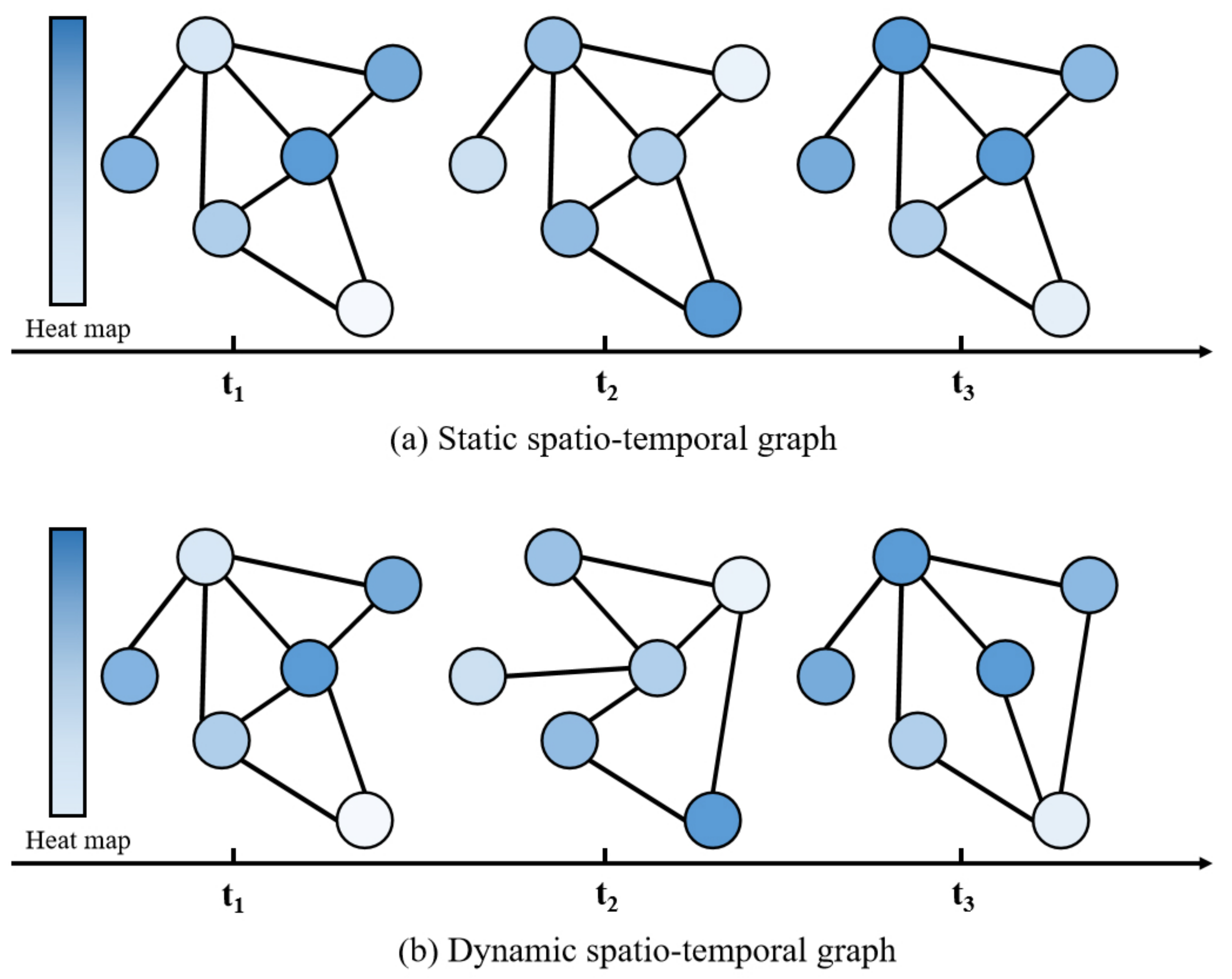}
\caption{The schematic diagram of static and dynamic spatio-temporal graphs. The color shades of the nodes represent the numerical differences in some predictable features.}
\label{fig:stg_type} % FIG
\vspace{-3mm}
\end{figure}

Suppose we obtain some observations from sensors, denoted as $\boldsymbol{X}=\{\mathcal X_{t}\in \mathbb{R}^{N\times F}|t=0,\dots,T\}$, where $N$ is the number of spatial vertices and $F$ is the number of features.
\textbf{Spatio-temporal Graph} is an efficient structure to characterize the relationships between different vertices in a certain spatial and temporal range. We can represent a spatio-temporal graph as $\mathcal{G}_{t}=(\mathcal{V}, \mathcal{E}_{t}, \boldsymbol{A}_{t})$, where $\mathcal{V}$ is the vertices set, $\mathcal{E}_{t}$ is the edge set, and $A_{t}$ denotes the adjacency matrix at time $t$. In most scenarios, the size of $\mathcal{V}$ is static, while the size of $\mathcal{E}_{t}$ can be time-varying or constant, which indicates that $\boldsymbol{A}_{t}\in \mathbb{R}^{N\times N}$ also changes with $\mathcal{E}_{t}$. In terms of connectivity, spatio-temporal graphs can be either directed or undirected, as well as weighted or unweighted. From the perspective of evolution, the structure of spatio-temporal graphs can be either static or dynamic. Figure~\ref{fig:stg_type} illustrates the difference between static and dynamic spatio-temporal graphs. The appropriate type of spatio-temporal graph to construct depends on the task and the given data conditions.

Generally, the construction methods of predefined spatio-temporal graphs in urban computing systems can be divided into four categories: topology-based, distance-based, similarity-based, and interaction-based. 

\textbf{Topology-based graph:} In the context of urban systems, topology-based graphs are usually constructed based on given topology structures, such as road networks~\cite{guo2021learning,wang2020traffic}. The adjacency matrix of a topology-based graph can be formulated as:
\begin{equation}
  a_{ij}^{t}=\left\{\begin{aligned}
  1, &\ \text{if } v_{i} \text{ connects to } v_{j}  \\ 
  0, &\ \text{otherwise}
  \end{aligned},
  \right.
\end{equation}
where $a_{ij}^{t}$ denotes an element in adjacency matrix at time $t$, $v_{i}$ and $v_{j}$ are different vertices in the graph. Since the connections in topology structures can be symmetrical or asymmetrical, the topology-based graphs can be directed or undirected. Topology only represents connections in non-Euclidean spaces, thus the topology-based graphs are unweighted. In addition, the topology structures in urban systems are usually fixed for quite a long time, so we can treat them as static graphs.     

\textbf{Distance-based graph:} According to first law of geography, \emph{i.e.}, ``Everything is related to everything else, but near things are more related to each other'', we can construct a distance-based graph when a predefined topology is absent. In most applications, the elements in the adjacency matrix are calculated using a kernel function that takes the distances into account~\cite{chai2018bike,yu2017spatio,jin2022deep}. Gaussian radial basis function and inverted function are two common kernel functions used in previous literature. For example, the adjacency matrix of a distance-based graph with Gaussian radial basis can be computed as:
\begin{equation}
  a_{ij}^t=\left\{\begin{aligned}
  &\frac{\exp (-\left\|d_{ij}^t\right\|_2)}{\sigma}, \text{if} \ d_{ij}^t < \epsilon \\ & \ \ \ \ \ \ 0 , \ \ \ \ \ \text{otherwise}
  \end{aligned},
  \right.
\end{equation}
where $d_{ij}^{t}$ denotes the distance between node $i$ and node $j$ at time $t$; $\epsilon$ is a predefined threshold to control the sparsity of the adjacency matrix; $\sigma$ is a hyper-parameter to control the distribution. 
\begin{figure*}[!t] 
\vspace{-2mm}
\centering
\includegraphics[width=0.9 \textwidth]{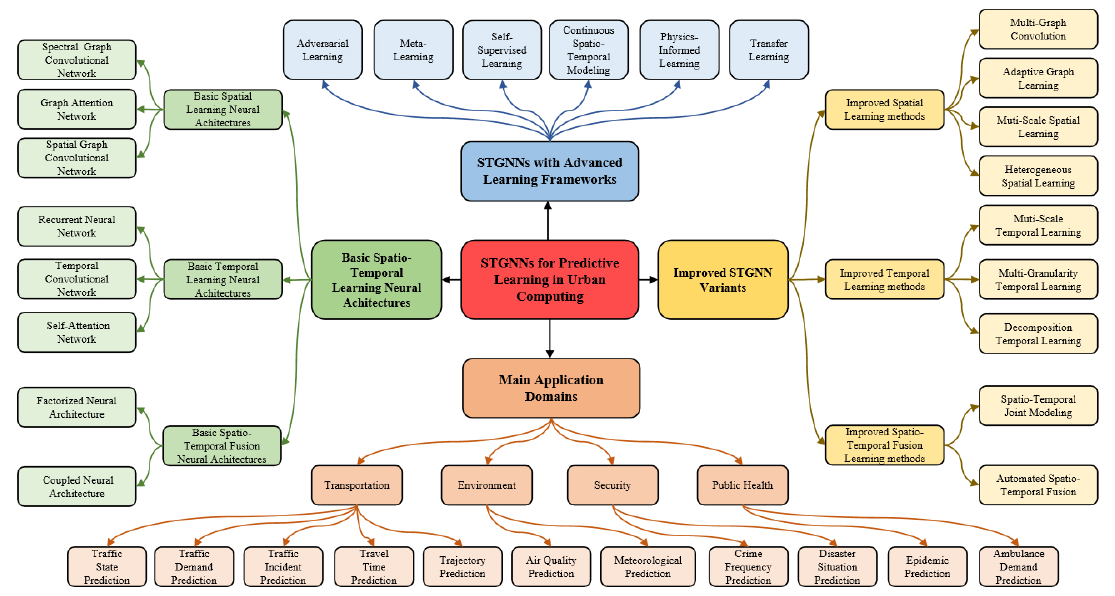}
\caption{The taxonomy for STGNN in our survey.}
\label{fig:taxonomy} % FIG
\vspace{-2mm}
\end{figure*}

\textbf{Similarity-based graph:} Similarity can provide insights into the relations between different entities from a semantic perspective. Similarity-based graphs can be constructed based on either the proximity of time series~\cite{liu2020physical,li2022automated,shi2020predicting} or similarity of the spatial attribute, \emph{e.g.}, Point of Interest (POI)~\cite{geng2019spatiotemporal}. In scenarios where additional data is unavailable, similarity-based graphs are typically constructed based on the similarity of time series. Pearson Correlation Coefficient (PCC) and Dynamic Time Wrapping (DTW) are two prevalent methods used to calculate the similarity between time series. For instance, the adjacency matrix of a similarity-based graph computed by PCC is defined as:
\begin{equation}
  a_{ij}^t=\left\{\begin{aligned}
  &\frac{\sum_{i=1}^n\left(x_i^{0:t}-\bar{x_i^{0:t}}\right)\left(x_j^{0:t}-\bar{x_j^{0:t}}\right)}{\sqrt{\sum_{i=1}^n\left(x_i^{0:t}-\bar{x_i^{0:t}}\right)^2} \sqrt{\sum_{i=1}^n\left(x_j^{0:t}-\bar{x_j^{0:t}}\right)^2}} \\ & \ \ \ \ \ \ \ \ \ \ \ \ \ \ 0 , \ \ \ \ \ \text{otherwise}
  \end{aligned},
  \right.
\end{equation}
where $x_i^{0:t}$ and $x_j^{0:t}$ represent the time series of node $i$ and node $j$ of a given time span $t$, respectively; $\bar{x_i^{0:t}}$ and $\bar{x_j^{0:t}}$ are the mean value of the time series of node $i$ and node $j$, and $n$ denotes the number of samples over the time span $t$.
%$\epsilon$ is a threshold to control the sparsity of adjacency matrix.

\textbf{Interaction-based graph:} The interaction between different locations can express their connection from the perspective of information flow~\cite{chai2018bike,jin2022deep}. This is especially important when representing the characteristics of mobility, as the proportion of flow between two nodes can indicate the strength of their connection. Hence, the adjacency matrix of an interaction-based graph can be written as:
\begin{equation}
  a_{ij}^t=\left\{\begin{aligned}
  &\frac{F_{ij}^{t}}{\sum_{m\in N(i)} F_{im}^{t}}, \ if \ F_{ij}^{t}>0 \\ & \ \ \ \ \ \ 0 , \ \ \ \ \ \text{otherwise}
  \end{aligned},
  \right.
\end{equation}
where $F_{ij}^{t}$ denotes the flow from node $i$ to node $j$ at time $t$; $N(i)$ indicates the set of nodes that interact with node $i$; $F_{im}^{t}$ is the flow from node $i$ to other nodes (\emph{e.g.}, $m$) in a set $N(i)$ at time $t$.

% As show in Table~\ref{tab:stg}, we sort out the differences in connectivity and evolutionary of different construction methods based on the existing literature.
% \begin{table}[!htb]
% %\small
% \caption{Spatio-temporal graph construction analysis from connectivity and evolutionary perspectives}
% %\vspace{-3mm}
% \centering
% \scalebox{0.75}{\begin{tabular}{|c|c|c|c|c|c|c|}
% \hline \multirow{2}{*}{ Construction methos} & \multicolumn{4}{|c|}{ Connectivity } & \multicolumn{2}{c|}{ Evolutionary } \\
% \cline { 2 - 7 } & Directed & Undirected & Weighted & unweighted & Static & Dynamic \\
% \hline Topology-based & $\sqrt{ }$ & $\sqrt{ }$ & & $\sqrt{ }$ & $\sqrt{ }$ & \\
% \hline Distance-based & & $\sqrt{ }$ & $\sqrt{ }$ & $\sqrt{ }$ & $\sqrt{ }$ & $\sqrt{ }$ \\
% \hline Similarity-based & & $\sqrt{ }$ & $\sqrt{ }$ & $\sqrt{ }$ & $\sqrt{ }$ & $\sqrt{ }$ \\
% \hline Interaction-based & $\sqrt{ }$ & $\sqrt{ }$ & $\sqrt{ }$ & $\sqrt{ }$ & $\sqrt{ }$ & $\sqrt{ }$ \\
% \hline
% \end{tabular}}
% \label{tab:stg}
% %\vspace{-2mm}
% \end{table}

% \begin{figure}[h] 
% \centering
% %\vspace{-2mm}
% \includegraphics[width=0.45 \textwidth]{f2.png}
% \caption{Types of spatio-temporal graph from three perspectives}
% \label{fig:stg_type} % FIG
% %\vspace{-3mm}
% \end{figure}

In addition to the common predefined graph construction methods mentioned above, many relations in urban systems are implicit and difficult to be directly predefined. Therefore, spatio-temporal graphs based on adaptive learning have been proposed in some recent works. More details about these methods can be found in Section~\ref{sec:adpative}.

\section{Taxonomy}~\label{sec:tax}
This section provides a taxonomy of STGNNs for predictive learning in urban computing, which is also a generalization of our follow-up content. As shown in Figure~\ref{fig:taxonomy}, there are four main parts in our survey that need to be highlighted: main application domains, basic spatio-temporal learning neural architecture, improved spatio-temporal learning methods, and advanced methods combined with STGNNs. We present an overview of specific predictive learning tasks based on major application domains in Section~\ref{sec:domain}. In Section~\ref{sec:basic_arch}, we review the fundamental neural architectures of STGNNs from three perspectives: spatial learning, temporal learning and spatio-temporal fusion.  Subsequently, in Section~\ref{sec:improved}, we examine the enhanced spatio-temporal dependency learning methods from the same perspectives as in Section~\ref{sec:basic_arch}. Finally, we discuss the advanced techniques combined with STGNNs in Section~\ref{sec:advanced}.

\begin{figure}[!b] 
\vspace{-3mm}
\centering
\includegraphics[width=0.4 \textwidth]{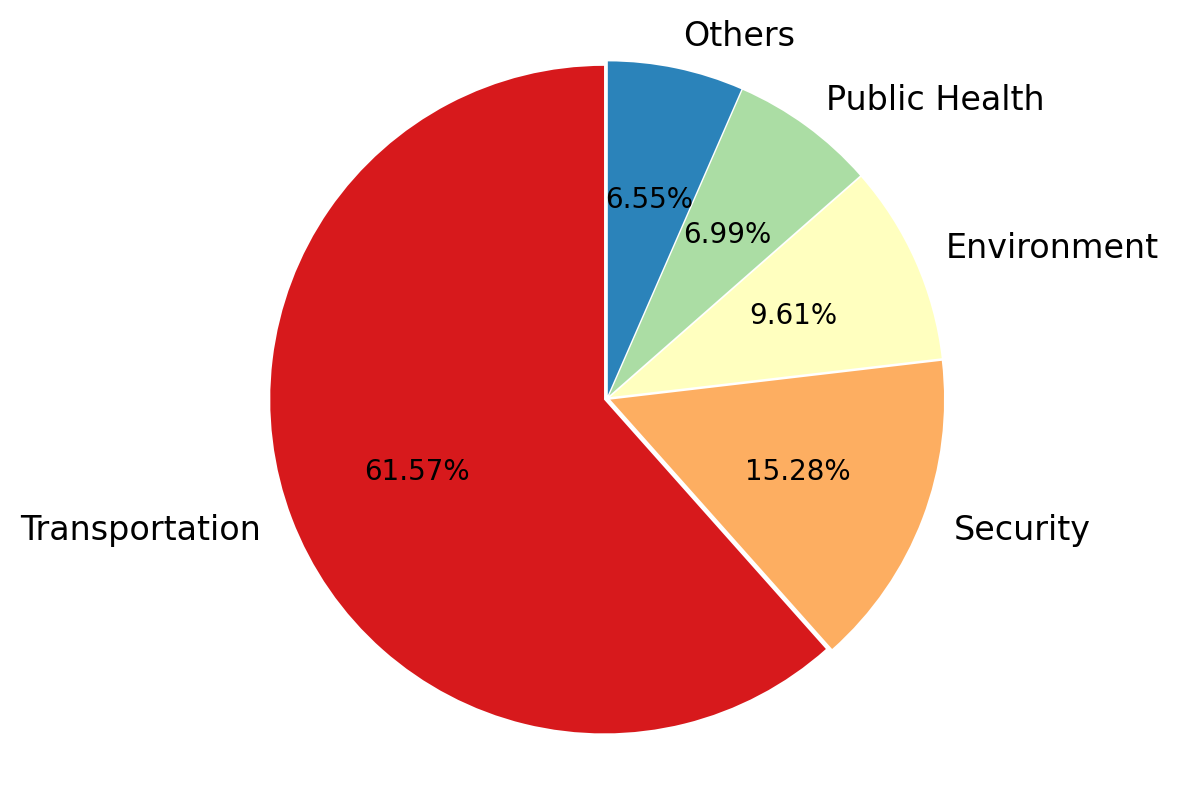}
\caption{The summary of the different application domains of STGNN in urban computing.}
\label{fig:statistical} % FIG
\vspace{-3mm}
\end{figure}

\section{Application Domains \& Task Description}\label{sec:domain}
This section delves into the primary application domains and specific predictive learning tasks in urban computing. Based on the available literature in recent years, we conducted a statistical analysis of the various application domains of STGNN in urban computing. Figure~\ref{fig:statistical} illustrates the main application domains of STGNN, which encompass transportation, safety, environment, and public health. Among these, transportation is the most widely studied application domain of STGNN, constituting over 60\% of the existing literature.

\subsection{Transportation}
Modern urban systems have numerous sensors distributed across traffic road networks and critical regions to monitor changing traffic states, such as flow and speed. The objective of traffic state prediction is to forecast future traffic states based on historical traffic states within a particular spatial range. 
% As shown in Figure~\ref{fig:tsp}, 
Traffic state prediction can be divided into two main categories: 
\begin{itemize}[leftmargin=*]
    \item \textbf{Network-based prediction.} The object of network-wide prediction is usually the traffic flow or speed on the given road networks~\cite{zhao2019t,guo2021learning,ji2022stden,li2017diffusion,yu2017spatio,zhang2019multistep}. The basic graph structures can be directly converted from road networks in most previous works. 

    \item \textbf{Region-based prediction.}  This task aims to forecast the traffic (\emph{e.g.}, crowd flow) in urban areas~\cite{wang2021gallat,zhang2021traffic,sun2020predicting,zhang2020spatial}. In this case, the whole urban area is partitioned into irregular or regular regions, and a spatio-temporal graph can be constructed based on the distances, connectivity, semantic correlations between different regions, etc. 
\end{itemize}

In general, traffic state prediction tasks can be summarized in the following form:
\begin{equation}
\left[\mathcal X_{\left(t-T^{\prime}+1\right)}, \cdots, \mathcal X_{(t)} ; \mathcal{G}\right] \stackrel{f(\cdot)}{\longrightarrow}\left[\mathcal X_{(t+1)}, \cdots, \mathcal X_{(t+T)}\right],
\end{equation}
where $\mathcal X_{(t)}\in \mathbb{R}^{N\times d}$ denotes the traffic states of $N$ vertices at time step $t$, $\mathcal{G}$ is the constructed graph structure, ${f(\cdot)}$ is the corresponding STGNN model for making predictions.
% \begin{figure}[!b] 
% \centering
% \vspace{-1em}
% \includegraphics[width=0.46 \textwidth]{f5.png}
% \caption{The two categories of traffic state prediction.}
% \label{fig:tsp} % FIG
% \end{figure}

\subsubsection{Traffic Demand Prediction}
Accurately predicting urban traffic demand patterns (\emph{e.g.}, taxi demands, rail transit passenger demands, and bike-sharing demands) in various regions can facilitate traffic scheduling to alleviate congestion during rush hours.
%This task aims to predict future traffic demand by historical traffic demand in a certain spatial range. 
Demands can be broadly categorized into three main types: origin demands, destination demands, and origin-destination (OD) demands. 
Predicting origin and destination demands is similar to region-based traffic state prediction, \emph{i.e.}, forecasting future demands based on historical demands in $N$ regions~\cite{jin2022deep,jin2020urban,geng2019spatiotemporal,ye2021coupled}. However, OD demand prediction is somewhat distinct, requiring prediction of future origin-destination matrices using historical OD matrices~\cite{wang2019origin,huang2023odformer,hu2020stochastic,dapeng2021dynamic,liu2022online,wang2021passenger}. To be specific, the outputs of OD demand prediction are a series of matrices with size $N\times N$, which can characterize the flow demand among these region pairs.

\subsubsection{Traffic Incident Prediction}
%Traffic state prediction tasks focus on the continuous numerical prediction of various spatio-temporal attributes of the road networks. 
With the dramatic increase in the number of vehicles, more and more traffic incidents such as congestion and accidents have occurred, placing significant pressure on urban traffic management. The aim of the traffic incident prediction task is to predict some important properties (\emph{e.g.}, occurrence probability, occurrence time) of these incidents that may occur on road networks~\cite{wang2021gsnet,yu2021deep,wang2022event,zhou2020riskoracle,jin2023npp}. In addition to differences in the objects being predicted, similar to the traffic state prediction task, accurate traffic incident prediction also requires capturing spatio-temporal dependencies on road networks by building STGNN models. 
% Compared with the relatively macroscopic traffic state prediction, the incident-oriented prediction can more accurately respond to various emergencies in the traffic system for early warning.

\subsubsection{Travel Time Prediction}
Travel time prediction is highly valued in industries, especially in online map navigation and ride-hailing software, which can significantly enhance the user experience. This task aims to predict the travel time of a given trajectory based on historical traffic states on road networks. To predict travel time more accurately, not only trajectory characteristics need to be considered, but also the spatio-temporal dynamics (\emph{e.g.}, flow, speed) attached to road networks should be addressed. Under this circumstance, spatio-temporal graphs are established based on road networks. So far, large technology companies such as Baidu~\cite{fang2020constgat,huang2022dueta}, Google~\cite{derrow2021eta}, and DiDi~\cite{fu2020compacteta} have developed practical travel time prediction functions on their online platforms. 
% STGNN-based travel time prediction can be defined as follows:
% \begin{equation}
% \mathcal{F}(P_{t}|X_{t-w : t},\mathcal{G})\rightarrow T_{g},T_{l}
% \end{equation}
% where $P_t$ denotes the given trajectory with departure time $t$, $X_{t-w : t}$ denotes the spatio-temporal features in historical time window $w$ attached to the given road network $\mathcal{G}$. $T_{g}$ and $T_{l}$ respectively represent the global travel time of the entire trajectory and local travel time of the road segment.

% \begin{figure}[h] 
% \centering
% %\vspace{-2mm}
% \includegraphics[width=0.45 \textwidth]{f5.png}
% \caption{The travel time prediction function in mobile application.}
% \label{fig:tte} % FIG
% %\vspace{-3mm}
% \end{figure}

\begin{table*}[h]
\centering
\caption{The public datasets for main application domains.}
\label{tab:dataset}
\resizebox{\textwidth}{!}{%
\begin{tabular}{|l|l|l|l|}
\hline
Domain                           & Dataset                 & Link                                                           & Reference \\ \hline
\multirow{13}{*}{Transportation} & California-PEMS        & http://pems.dot.ca.gov/                                        & \cite{guo2021learning,li2022automated,wu2022traversenet,li2021spatial,song2020spatial,jin2022automated}          \\ \cline{2-4} 
                                 & METR-LA                 & https://www.metro.net/                                         &  \cite{zheng2020gman,li2017diffusion,yu2017spatio,chen2020multi,wu2019graph,li2021dynamic,han2021dynamic}         \\ \cline{2-4} 
                             & NYC taxi                & https://www1.nyc.gov/site/tlc/about/tlc-trip-record-data.page  & \cite{ye2021coupled,zhang2020spatial,liu2022msdr,li2022lightweight,sun2020predicting}          \\ \cline{2-4} 
                                 & San Francisco taxi      & https://crawdad.org/ crawdad/epfl/mobility/20090224/           & \cite{zhao2019t,xie2020deep}          \\ \cline{2-4} 
                             & T-Drive Taxi & https://www.microsoft.com/en-us/research/publication/t-drive-trajectory-data-sample/ & \cite{bai2019spatio,ali2022exploiting}          \\ \cline{2-4}                                   
                                 & NYC bike                & https://www.citibikenyc.com/sytem-data                        &  \cite{zhou2021urban,li2022lightweight,chai2018bike,peng2021dynamic,zhang2021traffic,ali2022exploiting}         \\ \cline{2-4} 
                                 & Chicago bike            & https://www.divvybikes.com/system-data                         & \cite{chai2018bike,wang2021spatio,wang2022multivariate}          \\ \cline{2-4} 
                                 & NYC accident            & https://data.cityofnewyork.us/                                 &  \cite{zhou2020riskoracle,zhou2020foresee,wang2021gsnet,wang2021incident}         \\ \cline{2-4} 
                                 & Chicago accident        & https://data.cityofchicago.org/                                & \cite{wang2021gsnet,wu2020hierarchically}          \\ \cline{2-4} 
                                 & Chengdu taxi trajectory & http://www.dcjingsai.com./                                     & \cite{wang2021graphtte,wang2023multitask,jin2021spatio,jin2021hierarchical,jin2022stgnn}          \\ \cline{2-4} 
                                 & Porto taxi trajectory   & https://www.kaggle.com/crailtap/taxi-trajectory.               & \cite{jin2022stgnn,jin2021spatio,jin2021hierarchical,jiang2023self}          \\ \cline{2-4}                                
                             & ETH walking pedestrians & https://data.vision.ee.ethz.ch/cvl/aem/ewap\_dataset\_full.tgz & \cite{peng2021stirnet,shi2021sgcn,mohamed2020social,malla2021social,liu2022social,lian2022ptp}          \\ \cline{2-4} 
                                 & UCY walking pedestrians & https://graphics.cs.ucy.ac.cy/research/downloads/crowd-data    &  \cite{peng2021stirnet,shi2021sgcn,mohamed2020social,lian2022ptp}         \\ \hline
\multirow{5}{*}{Environment}     & Beijing air quality     & https://biendata.com/competition/kdd\_2018/data/               & \cite{lin2018exploiting,han2021joint}          \\ \cline{2-4} 
                                 & Shanghai air quality    & http://www.cnemc.cn/en/                                        & \cite{han2021joint}          \\ \cline{2-4} 
                                 & WeatherBench            & https://mediatum.ub.tum.de/1524895                             &\cite{lin2022conditional}           \\ \cline{2-4} 
                                 & Denmark wind speed      & https://sites.google.com/view/siamak-mehrkanoon/code-data      &\cite{stanczyk2021deep,rathore2021multi}           \\ \cline{2-4} 
                                 & Dutch wind speed        & https://github.com/HansBambel/multidim\_conv                   &\cite{stanczyk2021deep}           \\ \hline
\multirow{6}{*}{Public safety}        & NYC crime               & https://data.cityofnewyork.us/                                 &\cite{xia2021spatial,li2022spatial,sun2022spatial}            \\ \cline{2-4} 
                                 & Chicago crime           & https://data.cityofchicago.org/                                &\cite{xia2021spatial,tekin2022crime,li2022spatial,zhang2020graph,sun2022spatial}          \\ \cline{2-4} 
                                 & San Francisco crime     & https://datasf.org/openda                                      &\cite{wang2021spatio,jin2022adaptive}           \\ \cline{2-4} 
                                 & San Francisco fire      & https://datasf.org/openda                                      &\cite{wang2021spatio,jin2022adaptive}           \\ \cline{2-4} 
                                 & Japan typhoon           & http://agora.ex.nii.ac.jp/digital-typhoon/                     &\cite{yang2022spatio,zhou2020classification}           \\ \cline{2-4} 
                                 & California earthquake   & https://service.iris.edu/                                      &\cite{zhang2022spatiotemporal,feng2022gtrans}           \\ \hline
\multirow{4}{*}{Public health}   & US Covid-19             & https://github.com/CSSEGISandData/COVID-19                     &\cite{wang2022causalgnn,xie2022epignn,li2020study,yu2023spatio,kapoor2020examining}           \\ \cline{2-4} 
                                 & Italy Covid-19          & https://github.com/pcm-dpc/COVID-19                            &\cite{la2020epidemiological}           \\ \cline{2-4} 
                                 & Japan-Prefectures ILI   & https://tinyurl.com/y5dt7stm                                   &\cite{deng2019graph}        \\ \cline{2-4} 
                                 & US ILI                  & https://tinyurl.com/y39tog3h                                   &\cite{deng2019graph}           \\ \hline
\end{tabular}%
}
\end{table*}
\subsubsection{Trajectory Prediction}
Trajectory prediction is a crucial task for comprehending the intricate group dynamics of humans and vehicles~\cite{liu2022social,lu2022vehicle,peng2021stirnet,mohamed2020social,mo2022multi,mo2023map}, which fosters advancements in autonomous driving and urban monitoring technologies. There are some correlations or interactions in the movement patterns of agents in the group, thus we can build spatio-temporal graphs based on the relations between different agents within a group.  Upon creating these spatio-temporal graphs, STGNNs can be devised to predict the coordinates that agents may occupy in the future, considering their historical traversal coordinates, thereby facilitating the predictions of future trajectories.

% \begin{figure}[h] 
% \centering
% \vspace{-3mm}
% \includegraphics[width=0.45 \textwidth]{f4.png}
% \caption{An example of social interactions between different agents within a group~\cite{liu2022social}.}
% \label{fig:traj_p} % FIG
% \vspace{-3mm}
% \end{figure}

% \subsubsection{Other prediction tasks}
% Aside from the mainstream transportation application scenarios mentioned earlier, there are also some relatively niche scenarios that use STGNN techniques to improve prediction outcomes. For example, parking availability prediction~\cite{zhang2020semi,zhao2021mepark} and traffic delay prediction~\cite{zhang2021train,zhang2022interpretable} are two emerging tasks in transportation management that can be addressed with STGNN. Analogous to other prevalent research topics, these prediction tasks adopt STGNNs to better learn spatio-temporal contextual representations on traffic networks, yielding more accurate predictions. 

\subsection{Environment}

\subsubsection{Air Quality Prediction}
Air quality has become a pressing issue that needs immediate attention and improvement. Accurate air quality prediction can not only assist  governments in formulating energy-saving and emission-reduction policies but also provide guidance for residents' outdoor activities. Air quality index (AQI), PM2.5, and emissions are the indicators we are among the most significant indicators of concern. The related data are collected by city-level or national-level monitoring stations~\cite{liang2022airformer,han2021joint}. Due to the fluidity of the air, monitoring stations that are geospatially close or sharing the same wind direction may collect correlated results~\cite{zhou2021forecasting,wang2020pm2,liang2018geoman}. Hence, utilizing STGNN models can not only establish such spatial dependencies but also capture the time-varying dynamics of air quality. 

\subsubsection{Meteorological Prediction}
Meteorological forecasting is another research topic intimately connected to the environment and human society. Similar to air quality data, meteorological data are also collected by distributed monitoring stations. However, the correlations between different stations could be more complex and susceptible to a greater number of factors. In recent years, STGNN-based approaches have been progressively applied in various meteorological prediction scenarios such as temperature prediction~\cite{lin2022conditional,chen2022physics,jia2021physics}, frost prediction~\cite{lira2021frost} and wind prediction~\cite{rathore2021multi,khodayar2018spatio,stanczyk2021deep}, showcasing their superior performance in practice.

\subsection{Public Safety}

\subsubsection{Crime Frequency Prediction}
Effectively combating and preventing crime is the foundation for ensuring urban safety. Accurate prediction of crime frequency can assist governments in understanding real-time crime dynamics and allocating police resources rationally. Most existing work in this research line focus on crime frequency prediction in urban areas. Given that different urban regions have distinct functions, POI, and other characteristics, these factors could contribute to varying crime types and trends. However, regions with similar characteristics or close distances may exhibit latent correlations in crime incidents~\cite{xia2021spatial,qian2020gest,song2021dgcn}. Consequently, many previous studies~\cite{xia2021spatial,qian2020gest,jin2021gsen,feng2022gtrans,wang2018graph,zhang2020graph,sun2022spatial,wang2022hagen} have introduced a series of STGNN to capture these correlations to reduce the prediction errors.

\subsubsection{Disaster Situation Prediction}
Natural disasters, \emph{e.g.}, earthquakes, have posed big challenges to the safety of human society since ancient times. Disaster situation prediction can enable governments to implement disaster prevention measures, allocate disaster relief materials, and evacuate residents in a timely manner. STGNNs have been a fruitful approach in this task to model correlated and heterogeneous features across geographical locations. Currently, literature has introduced the STGNN models into scenarios such as flood prediction~\cite{feng2022graph,yuan2022spatio}, fire prediction~\cite{jin2022adaptive,jin2020ufsp}, typhoon forecasting~\cite{yang2022spatio,zhou2020classification,farahmand2021spatial} and earthquake prediction~\cite{dougan2022structural,mcbrearty2022earthquake,zhang2022spatiotemporal}.

\subsection{Public Health}
\subsubsection{Epidemic Prediction}
Epidemics are one of the greatest challenges to the public health systems, especially the novel coronavirus that has been prevalent in recent years, which has caused more than six million deaths worldwide. Therefore, accurately predicting the spread of epidemics is an important but challenging task, which can provide data support for the strengthening strategy of the urban public health systems. Some recent existing works have employed STGNN models to address the national-level~\cite{ngoc2021forecasting,kapoor2020examining,xie2022epignn,panagopoulos2021transfer,yu2023spatio,gao2021stan} or international-level~\cite{deng2020cola} epidemic prediction tasks. Many of them combine the mathematical formulations of epidemic dynamics and the modeling of spatio-temporal graphs, which have achieved better prediction results than traditional methods~\cite{sun2022using,zheng2020spatial,la2020epidemiological,gao2021stan}.

\subsubsection{Ambulance Demand Prediction}
In today's aging society, the allocation of ambulance resources is a challenging task that needs careful consideration. Accurate ambulance demand prediction can effectively alleviate the burden on the urban healthcare systems. Since there could be time-varying correlations in public medical resources, traffic conditions, and demand patterns among different regions of the social systems, STGNN-based methods have increasingly been exploited to learn these multi-view spatial correlations in recent years~\cite{wang2021forecasting,jin2021predicting,munasinghe2022using}. 

\subsection{Other Application Domains}
In addition to the four main application domains mentioned above, other scenarios where the spatio-temporal graph structures can be established based on the intrinsic relations of data are potential areas for the development of STGNN-based predictive learning models.
In recent years, STGNN-based predictive learning models have also been promoted to other domains such as energy, economy, finance, and production. In the energy domain, STGNN models have been utilized in wind power prediction~\cite{yu2022spatio,li2022spatiotemporal} and photovoltaic power prediction~\cite{simeunovic2021spatio}. In economy, a typical application is nation-level regional economy prediction, where researchers have explored the usage of STGNN models~\cite{hui2020predicting,xu2020attentional}. 
%In the finance domain, STGNN models have been widely applied for stock prediction~\cite{wang2021hierarchical,wang2022adaptive,tan2022finhgnn}.
% In the production domain, Fan et al. first adopted STGNN model in predicting crop yield~\cite{fan2022gnn}.

\subsection{Open Datasets and Benchmarks}
As depicted in Table~\ref{tab:dataset}, we have compiled a list of some of the most frequently used public datasets from previous works in the primary application domains, including their details such as source links and related publications. 
% For example, California-PEMS is a set of real-world traffic datasets used for traffic prediction tasks. These datasets are collected from the California Department of Transportation's Performance Measurement System (PeMS) and are commonly used to evaluate and benchmark various traffic prediction models. The PEMS datasets are named according to the number of sensor stations in each dataset, such as PEMS-04 and PEMS-08. 
These datasets are widely used in the research field of transportation and urban mobility, particularly for traffic state prediction. Due to their high granularity, realistic nature, and real-world applicability, they serve as valuable resources for researchers working on spatio-temporal forecasting and traffic modeling.

There are also some well-known benchmarks in these primary application domains, especially in traffic prediction, such as BasicTS\footnote{https://github.com/zezhishao/BasicTS}, Traffic-Benchmark\footnote{https://github.com/tsinghua-fib-lab/Traffic-Benchmark}, DL-Traff\footnote{https://github.com/deepkashiwa20} and LargeST\footnote{https://github.com/liuxu77/LargeST}. For other application domains, there are fewer benchmarks, but there are some as follows, e.g., weatherbench2\footnote{https://github.com/google-research/weatherbench2} for meteorological forecasting and PM2.5-GNN\footnote{https://github.com/shuowang-ai/PM2.5-GNN} for air quality forecasting.

% \vspace{14em}

\section{Basic Neural Architectures}\label{sec:basic_arch}

\begin{figure}[!t] 
\centering
\vspace{-3mm}
\includegraphics[width=0.47 \textwidth]{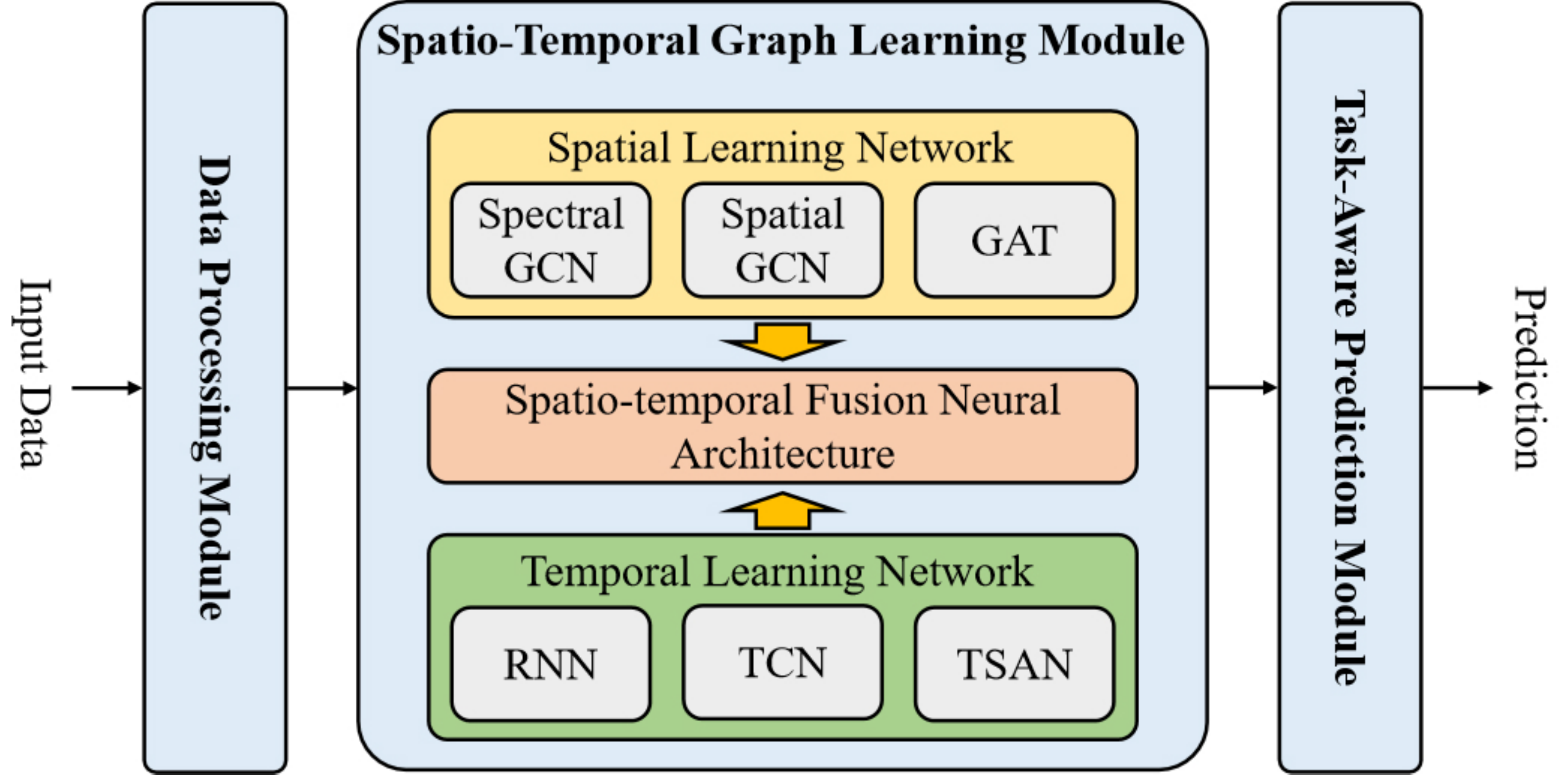}
\caption{The basic framework of STGNNs.}
% \caption{The basic framework of STGNNs for predictive learning in urban computing. It is composed of three major components: Data Processing Module (DPM), Spatio-Temporal Graph Learning Module (STGLM), and Task-aware Prediction Module (TPM).}
\label{fig:stgnn_framework} % FIG
\vspace{-3mm}
\end{figure}

Here we introduce basic neural architectures for STGNNs. As shown in Figure~\ref{fig:stgnn_framework}, the basic framework of STGNNs for predictive learning contains three main modules: Data Processing Module (DPM), Spatio-Temporal Graph Learning Module (STGLM), and Task-Aware Prediction Module (TPM). For predictive learning tasks in urban computing, DPM is responsible for constructing the spatio-temporal graph data from the raw data; STGLM seeks to capture hidden spatio-temporal dependencies from complex social systems, while TPM aims to map the spatio-temporal hidden representations from STGLM into the space of downstream prediction tasks. 
STGLM serves as the most vital component of STGNNs, which usually combines spatial learning networks and temporal learning networks organically through a certain spatio-temporal fusion method. 
Spatial learning networks may utilize spectral graph convolutional networks (Spectral GCNs)~\cite{defferrard2016convolutional}, spatial graph convolutional networks (Spatial GCNs)~\cite{kipf2016semi,hamilton2017inductive}, and graph attention networks (GATs)~\cite{velivckovic2017graph} as potential options.
Temporal learning networks, on the other hand, may incorporate recurrent neural networks (RNNs), temporal convolutional networks (TCNs), or temporal self-attention networks (TSANs). 
Compared with STGLM, TPM is a relatively simple neural network, thus the majority of existing research focuses on the design of the neural architectures in STGLM. 

\vspace{-1em}
\subsection{Graph Neural Networks}
Graph neural networks (GNNs) are fruitful tools for learning spatial dependencies in non-Euclidean space. In recent years, popular GNNs can be divided into three categories: spectral GCNs, spatial GCNs and GATs. 

\subsubsection{Spectral Graph Convolutional Network}
Initially, most GNNs were based on the Fourier transform, which converts the graph signal in the spatial domain into the spectral domain to conduct convolution calculations~\cite{ZHOU202057}.
% In this approach, Graph Fourier transform and Inverse Graph Fourier transform are performed to achieve the transformation between the two domains:
% \begin{equation}
% \mathcal{F}(\boldsymbol{x})=\boldsymbol{U}^T \boldsymbol{x}, \quad
% \mathcal{F}^{-1}(\boldsymbol{x})=\boldsymbol{U} \boldsymbol{x},
% \label{eq:s_gnn}
% \end{equation}
% where $\mathbf{U}$ denotes the matrix of eigenvectors of the normalized graph Laplacian.  Based on this, the graph convolution operation is defined as:
% \begin{equation}
% \begin{aligned}
% & \boldsymbol{g} \star \boldsymbol{x}=\mathcal{F}^{-1}(\mathcal{F}(\boldsymbol{g}) \odot \mathcal{F}(\boldsymbol{x})) =\boldsymbol{U}\left(\boldsymbol{U}^T \boldsymbol{g} \odot \boldsymbol{U}^T \boldsymbol{x}\right),
% \end{aligned}
% \label{eq:s_gnn2}
% \end{equation}
The notation $\odot$ is the convolution operator, $\mathbf{U}$ denotes the matrix of eigenvectors of the normalized graph Laplacian and $\mathbf{U}^T \mathbf{g}$ is the filter in the spectral domain. The graph convolution operation is defined as:
% Eq.~\ref{eq:s_gnn2} can be further simplified as:
\begin{equation}
\boldsymbol{g}_w \star \boldsymbol{x}=\boldsymbol{U g}_w \boldsymbol{U}^T \boldsymbol{x}.
\end{equation}
Most of the subsequent GNNs based on the spectral domain mainly improve the calculation method of $\boldsymbol{g}_w$. For example, ChebNet~\cite{defferrard2016convolutional} is one of the most popular Spectral GNN methods. 
According to the theory that $\boldsymbol{g}_w$ can be approximated by a truncated expansion of Chebyshev polynomials~\cite{hammond2011wavelets}.
% \cite{defferrard2016convolutional} further proposes ChebNet, which can be formulated as:
% \begin{equation}
% \begin{aligned}
% &\widetilde{\boldsymbol{L}}=\frac{2}{\lambda_{\max }} \boldsymbol{L}-\boldsymbol{I}_N,\\
% &\boldsymbol{g}_w \star \boldsymbol{x} = \sum_{k=0}^K w_k \boldsymbol{T}_k(\widetilde{\boldsymbol{L}}) \boldsymbol{x},
% \end{aligned}
% \end{equation}
% where $\widetilde{\boldsymbol{L}}$ is the normalized graph Laplacian, $\lambda_{\max }$ is the largest eigenvalue of $\boldsymbol{L}$, $\boldsymbol{T}_k(x)$ denotes the Chebyshev polynomials up to $k_{th}$ order, $w_k$ denotes a vector of Chebyshev coefficients. ChebNet eliminates the necessity of computing the eigenvectors of the Laplacian by employing the K-localized graph convolution.

\subsubsection{Spatial Graph Convolutional Network}
While spectral graph convolutional networks (GCNs) have made significant advancements, their primary limitation lies in their dependence on the graph Laplacian matrix. Whenever there is a change in the underlying graph structure, the graph Laplacian matrix must be recomputed, rendering spectral GCNs better suited to scenarios where the graph structure remains constant. 
To overcome the dependency on the graph Laplacian matrix, Kipf et al. simplify the graph convolution operation~\cite{kipf2016semi} by performing message passing in the spatial domain. We called this new form as spatial GCNs, which is defined as:
\begin{equation}
\boldsymbol{g}_w \star \boldsymbol{x} = w\left(\boldsymbol{I}_N+\boldsymbol{D}^{-\frac{1}{2}} \boldsymbol{A} \boldsymbol{D}^{-\frac{1}{2}}\right) \boldsymbol{x},
\end{equation}
where $\boldsymbol{A}$ is the adjacency matrix; $\boldsymbol{D}$ is the degree matrix; $w$ is the learnable parameters in the spatial GCN.

% However, the above spatial GCN has to take the full graph as input, which makes it hard to scale up to large graphs in practice. To address this problem, GraphSAGE~\cite{hamilton2017inductive} proposes a sampling aggregation approach to achieve flexible inductive learning on large graphs. 
% The aggregation operator in GraphSAGE is formulated as:
% \begin{equation}
% \begin{aligned}
% & \boldsymbol{h}_{\mathcal{N}(u)}^k \leftarrow \operatorname{Aggregate}_k\left(\left\{\boldsymbol{h}_{u^{\prime}}^{k-1}, \forall u^{\prime} \in \mathcal{N}_k(u)\right\}\right), \\
% & \boldsymbol{h}_u^k \leftarrow \sigma\left(\boldsymbol{W}^k \cdot \operatorname{Concat}\left(\boldsymbol{h}_u^{k-1}, \boldsymbol{h}_{\mathcal{N}(u)}^k\right)\right),
% \end{aligned}
% \end{equation}
% where $\mathcal{N}_k(u)$ denotes the set of neighbor nodes of $u$; $\mathbf{h}_{\mathcal{N}(u)}^k$ means the embedding of node $u$ after aggregation operation.

\subsubsection{Graph Attention Network}
To account for the importance of neighbor nodes in learning spatial dependencies, GAT~\cite{velivckovic2017graph} integrates the attention mechanism into the node aggregation operation as:
\begin{equation}
\begin{gathered}
\boldsymbol{h}_v^{t+1}=\rho\left(\sum_{u \in \mathcal{N}_v} \alpha_{v u} \boldsymbol{W h}_u^t\right), \\
\alpha_{v u}=\frac{\exp \left(\operatorname{LeakyReLU}\left(\boldsymbol{a}^T\left[\boldsymbol{W h}_v \| \boldsymbol{W h}_u\right]\right)\right)}{\sum_{k \in \mathcal{N}_u} \exp \left(\operatorname{LeakyReLU}\left(\boldsymbol{a}^T\left[\boldsymbol{W} \boldsymbol{h}_v \| \boldsymbol{W h}_k\right]\right)\right)},
\end{gathered}
\end{equation}
where $\alpha_{v u}$ denotes the attention scores of neighbor node $u$ to the central node $v$, $W$ is the weight matrix associated with the linear transformation for each node, and $\mathbf{a}$ is the weight parameter for attention output.
% To further stabilize the computational process of attention, the multi-head trick form also be introduced in GAT:
% \begin{equation}
% \begin{aligned}
% & \boldsymbol{h}_v^{t+1}=\|_{k=1}^K \sigma\left(\sum_{u \in \mathcal{N}_v} \alpha_{v u}^k \boldsymbol{W}_k \boldsymbol{h}_u^t\right), \\
% & \boldsymbol{h}_v^{t+1}=\sigma\left(\frac{1}{K} \sum_{k=1}^K \sum_{u \in \mathcal{N}_v} \alpha_{v u}^k \boldsymbol{W}_k \boldsymbol{h}_u^t\right),
% \label{eq:multi-attention}
% \end{aligned}
% \end{equation}
% where $\alpha_{v u}^k$ is the normalized attention score computed by the $k_{th}$ attention head. The aggregation method for multiple attention heads can be achieved by concatenation or an averaging operation. 

\subsection{Recurrent Neural Networks}
Recurrent neural networks (RNNs) are a class of deep neural networks for sequential learning based on recursive computations, and have found extensive applications in time series modeling. However, the vanilla version of RNNs is subject to a significant limitation -- the gradient vanishing or explosion problem during the training process~\cite{lukovsevivcius2009reservoir}. In response to this challenge, two of the most prominent variants of RNNs, \emph{i.e.}, Long Short-Term Memory (LSTM)~\cite{hochreiter1997long} and Gated Recurrent Units (GRU)~\cite{dey2017gate}, have been proposed. So far, GRU is the most widely used variant because it takes into account both high performance and low computational complexity.

GRU only has two efficient gated computational units: an update gate and a reset gate. $\mathbf{u}_{t}$ represents the update gate, which determines how to combine the information of the new input time step with the memory of the previous time step. $\mathbf{r}_{t}$ represents the reset gate, which defines the amount of memory reserved from the previous time step to the current time step. Although the learnable parameters of GRU are streamlined, its performance can be compared with LSTM in previous works, while improving the training and inference efficiency. The calculation process of GRU is defined as follows:
  \begin{equation}
	\begin{split}
		& \boldsymbol{u}_t =\sigma(\boldsymbol{W}_{u}\cdot x_t + \boldsymbol{U}_{u}\cdot \boldsymbol{C}_{t-1} + \boldsymbol{b}_{u}),\\
		& \boldsymbol{r}_t =\sigma(\boldsymbol{W}_{r}\cdot x_t+ \boldsymbol{U}_{r}\cdot \boldsymbol{C}_{t-1} + \boldsymbol{b}_{r}), \\
		&  \tilde{\boldsymbol{C}}_{t} ={\rm tanh}(\boldsymbol{W}_{C}\cdot x_t + \boldsymbol{U}_{C}(\boldsymbol{r}_t \odot \boldsymbol{C}_{t-1})+\boldsymbol{b}_{C}), \\
		&  \boldsymbol{C}_t=\boldsymbol{u}_t\odot \boldsymbol{C}_{t-1} + (1-\boldsymbol{u}_t)\odot \tilde{\boldsymbol{C}}_{t}.
  \label{eq:gru}
	\end{split}
\end{equation}
where $\tilde{\mathbf{C}}_{t}$ represents the candidate state of the current GRU unit after the calculation of the reset gate, $\mathbf{C}_{t}$ represents the state of the GRU unit after the calculation of the update gate.

\subsection{Temporal Convolutional Networks}
RNNs have been extensively applied for temporal learning in many spatio-temporal tasks, but their disadvantage is readily apparent: the recurrent structures necessitate the computation of sequences at every time step, leading to a substantial increase in computational cost and a consequent decrease in model efficiency. In contrast, Temporal Convolutional Networks (TCN) with their parallel 1D-CNN structures can address this problem effectively. 
% Similar to 2D-CNN applied in image recognition, 1D-CNN also operates and aggregates features through convolution kernels. The major difference is that the convolution kernel of 1D-CNN is one-dimensional and only slides on the time axis.

\subsubsection{Gated Temporal Convolutional Network}
Inspired by the gated mechanism in LSTMs and GRUs, we can also integrate it with pure 1D-CNN architecture to enhance the capability of temporal learning. We called this hybrid neural architecture a gated temporal convolutional network (Gated-TCN)~\cite{dauphin2017language}. The calculation process of Gated-TCN is defined as follows:
\begin{equation}
F(x) = tanh(\boldsymbol{\Theta_1}\star x )\odot\sigma(\boldsymbol{\Theta_2}\star x),
\end{equation}
where $\mathbf{\Theta_1}$ and $\mathbf{\Theta_2}$ represent the learnable parameters of the convolution kernel in two different 1D-CNNs, respectively; $\star$ denotes the convolution operation; $\odot$ is the element-wise multiplication mechanism; $\sigma(\mathbf{\Theta_2}\star x)$ indicates the gating unit, which is utilized to control the utilization rate of historical information.

\subsubsection{Causal Temporal Convolutional Network}
% Although 1D-CNN is an efficient parallel neural architecture, it fails to adequately model causal correlations in temporal learning. Usually, in the traditional neural networks, the connection of neurons in each layer is in the form of fully connected. It is not difficult to find that the fully connection violates the basic constraint of time series, because the output of the front (the previous time steps) neurons are connected to the input neurons later (the future time steps), which should not be allowed.
% Hence, we can employ a mask mechanism to partially remove the layer-by-layer links in the networks, and keep those links from the previous time steps to the future time steps, so that the network meets the principle of temporal dependencies. Meanwhile, in order to capture longer-range temporal dependencies more effectively, the 1D-CNN with the dilated factors~\cite{yu2015multi} increasing layer by layer has the capability to learn temporal dependencies from short-range to long-range, as shown in the Figure~\ref{fig:tcn}.
% The 1D-CNN with dilated factors is expressed as:

% \begin{figure}[!b] 
% \centering
% \includegraphics[width=0.48 \textwidth]{f7.png}
% \caption{The overview of causal temporal convolutional network with exponentially increasing dilated factors~\cite{oord2016wavenet}.}
% \label{fig:tcn} % FIG
% \end{figure}

While TCN is an efficient parallel neural architecture for sequential learning, it violates the temporal order of spatio-temporal graph data. Compared to standard TCNs, Causal TCNs, which are proposed in Wavenet~\cite{oord2016wavenet}, offer the additional benefit of explicitly modeling the causal nature of temporal data. This is achieved by removing connections between future time steps and past time steps, which eliminates the possibility of data leakage from future time steps to past time steps.
%, as depicted in Figure~\ref{fig:tcn}. 
Furthermore, in order to more effectively capture longer-range temporal dependencies, 1D-CNN with dilated factors~\cite{yu2015multi} can be used. By increasing the dilated factors layer by layer, this model has the capacity to learn temporal dependencies from a short range to a long range. A Causal TCN with dilated factors can be expressed as:
\begin{equation}
F(s)=\left(\boldsymbol{x} *_d f\right)(s)=\sum_i^{k-1} f(i) \cdot x_{s-d \cdot i},
\end{equation}
where $s$ is the input time series; $d$ represents the dilation factor, and the ordinary convolution operator is a special case of the dilated convolution operator when $d=1$; $s-d \cdot i$ refers to the positioning of certain historical information.

\subsection{Temporal Self-Attention Networks}
Self-attention networks represent a highly effective approach for capturing long-range temporal relationships among different time steps, with the most prominent example being the Transformer model~\cite{vaswani2017attention}. The Transformer comprises three primary components: a scaled dot-product attention network, a feed-forward network, and position encodings.
% \begin{figure}[!h] 
% \centering
% \includegraphics[width=0.46 \textwidth]{f8.png}
% \caption{The overview of Transformer~\cite{vaswani2017attention}}
% \label{fig:san} % FIG
% \end{figure}
The scaled dot-product network is the core part of Transformers, in which the attention calculation is formulated as:
\begin{equation}
\operatorname{Attention}(\boldsymbol{Q}, \boldsymbol{K}, \boldsymbol{V})=\operatorname{Softmax}\left(\frac{\boldsymbol{Q} \boldsymbol{K}^T}{\sqrt{\boldsymbol{d}_k}}\right) \boldsymbol{V},
\end{equation}
where queries $\boldsymbol{Q}$, keys $\boldsymbol{K}$, values $\boldsymbol{V}$ are three basic elements in the self-attention mechanism, which are obtained by non-shared linear transformations from the original input. $\boldsymbol{d}_k$ denotes the scaling factor, whose value is equal to the dimension of the model. 
% To stabilize the training process, this part can also employ the multi-head attention form, which is similar to Eq.~\ref{eq:multi-attention}.
Since Transformer contains no recurrence or convolution operator, we have to inject some positional information (eg., trigonometric function-based encoding) about the tokens in the sequence to consider the order of the sequence. 
%Trigonometric function-based encoding is a common positional encoding approach. 
% which is defined as:
% \begin{equation}
% \begin{aligned}
% P E_{(p o s, 2 i)} & =\sin \left(p o s / 10000^{2 i / d_{\text {model }}}\right), \\
% P E_{(p o s, 2 i+1)} & =\cos \left(p o s / 10000^{2 i / d_{\text {model }}}\right),
% \end{aligned}
% \end{equation}
% where $pos$ is the position and $i$ is the dimension. This method performs sine encoding and cosine encoding on even and odd positions respectively, to distinguish different positions. In addition to Trigonometric functions, we can leverage fully learnable position encodings to encode the position information.

\subsection{Spatio-Temporal Fusion Neural Architecture}\label{sec:stfa}
In addition to spatial learning networks and temporal learning networks, spatio-temporal fusion neural architecture represents another critical area, as it determines how spatial learning networks and temporal learning networks are integrated into the complete STGNN. Existing fusion neural architectures can be divided into two categories -- factorized or coupled neural architecture.

%\subsubsection{Stacked Neural Architecture}
\subsubsection{Factorized Neural Architecture}
\begin{figure}[h] 
\centering
\vspace{-3mm}
\includegraphics[width=0.45 \textwidth]{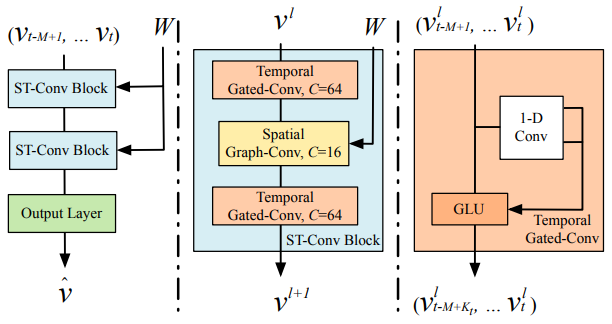}
\caption{The overview\protect\footnotemark of STGCN~\cite{yu2017spatio}.}
\label{fig:stgcn} % FIG
% \vspace{-3mm}
\end{figure}

\begin{figure}[h] 
\centering
% \vspace{-3mm}
\includegraphics[width=0.45 \textwidth]{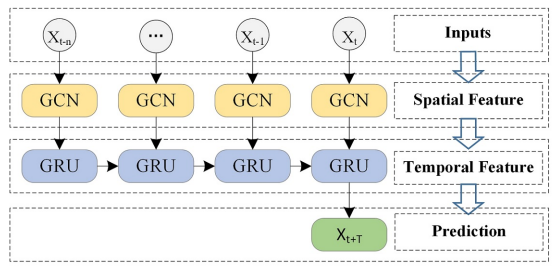}
\caption{The overview of T-GCN~\cite{zhao2019t}.}
\label{fig:tgcn} % FIG
\vspace{-1em}
\end{figure}

In factorized neural architectures, spatial learning networks and temporal learning networks are stacked in parallel or serially like building blocks layer by layer. There are two typical examples for factorized neural architectures in STGNN models, as shown in Figure~\ref{fig:stgcn} and Figure~\ref{fig:tgcn}, respectively. The first example is STGCN~\cite{yu2017spatio}, whose temporal learning network is TCN. In each ST-Conv block of STGCN, two TCNs and one GCN are stacked in series, forming a sandwich structure. Since this model learns temporal information through convolutional structures, its spatio-temporal learning method is parallelized, \emph{i.e.}, it receives all information of a given time window length as input at the same time. Mathematically, the calculation of each ST-Conv block in this model can be defined as follows::
\begin{equation}
v^{l+1}=\boldsymbol{\Gamma_1^l} *_{\mathcal{T}} \operatorname{ReLU}\left(\boldsymbol{\Theta^l} * _{\mathcal{G}}\left(\Gamma_0^l * \mathcal{T} v^l\right)\right),
\end{equation}
where $\Gamma_0^l$ and $\Gamma_1^l$ denote the upper and lower temporal convolutional kernel within block $l$, and $\Theta^l$ is the spectral kernel of graph convolution.

\footnotetext{The figures used in this paper but are not drawn by us (including Fig.~\ref{fig:stgcn} to Fig~\ref{fig:physical}) are all with the permission of their authors}

The second one is T-GCN~\cite{zhao2019t}, which utilizes GRUs for temporal learning. This model captures the spatio-temporal dependencies in a recursive manner. For each time step, graph signals are sequentially processed by GCN and GRU to learn spatial and temporal dependencies separately. The whole process of each stacked GCN and GRU in this model can be expressed as:
\begin{equation}
\begin{aligned}
&\boldsymbol{f}(\boldsymbol{X}, \boldsymbol{A})=\sigma\left(\boldsymbol{A} \boldsymbol{X} \boldsymbol{W}_0\right), \\
&\boldsymbol{u}_t=\sigma\left(\boldsymbol{W}_u\left[\boldsymbol{f}\left(\boldsymbol{A}, {\mathcal X}_{t}\right), \boldsymbol{H}_{t-1}\right]+\boldsymbol{b}_u\right), \\
&\boldsymbol{r}_t=\sigma\left(\boldsymbol{W}_r\left[\boldsymbol{f}\left(\boldsymbol{A}, {\mathcal X}_{t}\right), \boldsymbol{H}_{t-1}\right]+\boldsymbol{b}_r\right), \\
&\boldsymbol{c}_t=\tanh \left(\boldsymbol{W}_c\left[\boldsymbol{f}\left(\boldsymbol{A}, {\mathcal X}_{t}\right),\left(\boldsymbol{r}_t * \boldsymbol{H}_{t-1}\right)\right]+\boldsymbol{b}_c\right), \\
&\boldsymbol{H}_t=\boldsymbol{u}_t * \boldsymbol{H}_{t-1}+\left(1-\boldsymbol{u}_t\right) * \boldsymbol{c}_t,
\end{aligned}
\end{equation}
where $f(A,{\mathcal X}_{t})$ denotes the output of spatial GCN at time step $t$. Then $f(A,{\mathcal X}_{t})$ is put forward into GRU to obtain the hidden state at $t$.

\subsubsection{Coupled Neural Architecture}
\begin{figure}[h] 
\centering
% \vspace{-3mm}
\includegraphics[width=0.48 \textwidth]{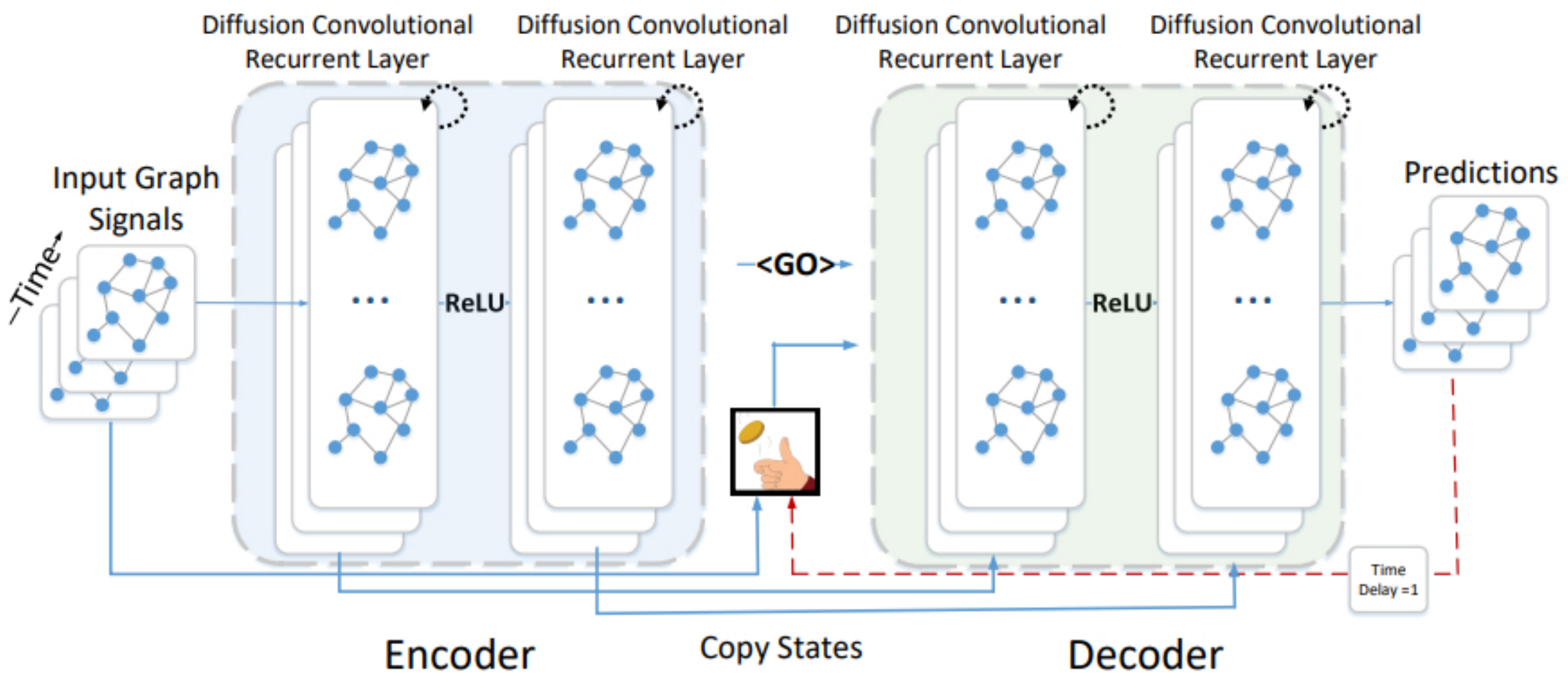}
\caption{The overview of DCRNN~\cite{li2017diffusion}.}
\label{fig:dcrnn} % FIG
% \vspace{-3mm}
\end{figure}

In coupled neural architectures, spatial learning networks are usually integrated into the architecture of temporal learning networks as embedded components. In STGNN, this type of neural architecture occurs almost exclusively in combinations of GNN-based spatial learning networks and RNN-based temporal learning networks. One example of a coupled neural architecture in STGNNs is the DCRNN~\cite{li2017diffusion}, which integrates GCN into the architecture of GRU, as illustrated in Figure~\ref{fig:dcrnn}. In this model, the original linear units in LSTM are replaced with a graph convolution operator, which can be written as:
\begin{equation}
\begin{aligned}
&\boldsymbol{r}_{t}=\sigma\left(\boldsymbol{\Theta}_r \star \mathcal{G}\left[{\mathcal X}_{t}, \boldsymbol{H}_{t-1}\right]+\boldsymbol{b}_r\right), \\
&\boldsymbol{u}_{t}=\sigma\left(\boldsymbol{\Theta}_u \star \mathcal{G}\left[{\mathcal X}_{t}, \boldsymbol{H}_{t-1}\right]+\boldsymbol{b}_u\right), \\
&\boldsymbol{C}_{t}=\tanh \left(\boldsymbol{\Theta}_C \star \mathcal{G}\left[{\mathcal X}_{t},\left(\boldsymbol{r}_{t} \odot \boldsymbol{H}_{t-1}\right)\right]+\boldsymbol{b}_c\right), \\
&\boldsymbol{H}_{t}=\boldsymbol{u}_{t} \odot \boldsymbol{H}_{t-1}+\left(1-\boldsymbol{u}_{t}\right) \odot \boldsymbol{C}_{t},
\end{aligned}
\end{equation}
where $\boldsymbol{\Theta}_r \star \mathcal{G}$ denotes the graph convolution operator with parameter $\boldsymbol{\Theta}_r$. Compared with the equation~\ref{eq:gru} of the original GRU, we can find that except for the internal graph convolution operator, the external calculation methods of the recurrent network are not much different. Similar to some neural translation models~\cite{cho2014learning}, DCRNN can also employ sequence-to-sequence structure to improve predictions. 

\section{STGNN Variants}~\label{sec:improved}
In Section~\ref{sec:basic_arch}, we have introduced the basic neural architectures of STGNNs, thereby augmenting the comprehension of the spatio-temporal learning paradigm within this research domain. However, in recent years, there have been numerous innovative methods devised to enhance the learning of spatio-temporal dependencies. In this section, we elaborate on some advanced STGNN variants that can better capture spatio-temporal dependencies for predictive learning in urban computing.

\subsection{Spatial Learning Methods}
\begin{figure}[!b] 
\centering
% \vspace{-3mm}
\includegraphics[width=0.48 \textwidth]{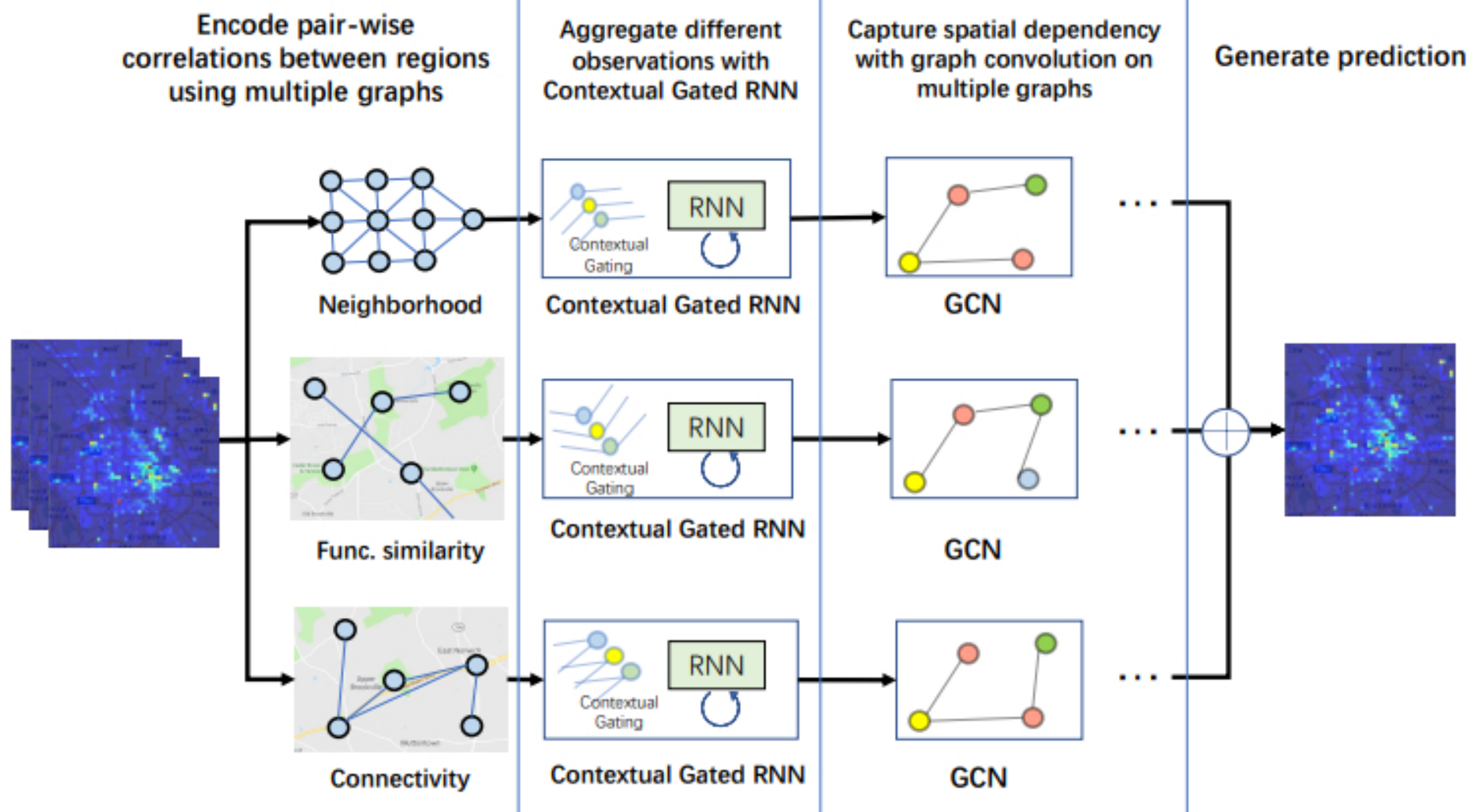}
\caption{The overview of STMGCN~\cite{geng2019spatiotemporal}.}
\label{fig:multi_gnn} % FIG
% \vspace{-3mm}
\end{figure}

\subsubsection{Multi-Graph Convolution}
In urban systems, there are often multiple types of spatial relations that exist simultaneously. For instance, in transportation systems, adjacent regions and regions with similar POIs may exhibit similar traffic patterns. Hence, jointly considering multiple spatial relations is necessary for spatio-temporal learning in STGNN. In recent years, a series of STGNN variants that integrate multi-graph convolutions have been proposed to address this challenge~\cite{liu2020physical,ni2022stgmn,he2022multi,jin2022deep,jin2020urban,geng2019spatiotemporal,chai2018bike,xu2020spatiotemporal}. Among them, STMGCN~\cite{geng2019spatiotemporal} is a typical model for urban ride-haling demand prediction, as shown in Figure~\ref{fig:multi_gnn}. 
This model first constructs multi-graph based on neighborhood, function similarity, and connectivity to characterize multiple spatial correlations. For each graph, contextual gated RNN and ChebNet are respectively adopted to capture temporal and spatial dependencies. Finally, the final prediction is obtained by fusing the parallelized multi-graph spatio-temporal hidden information.

\subsubsection{Adaptive Graph Learning}~\label{sec:adpative}
Despite its capability to capture multiple spatial correlations to some extent, multi-graph modeling still suffers from two limitations.  Firstly, the graph construction process may be insufficient and fail to account for other implicit correlations. Secondly, the rationality of graph construction may be questioned, particularly in the absence of sufficient domain knowledge to support it. To overcome these challenges, adaptive graph learning methods have been developed gradually. According to existing literature, adaptive graph learning methods in STGNN can be broadly categorized into two main categories: random initialization-based and feature initialization-based approaches.

\textbf{Random initialization-based} methods perform adaptive graph structure learning via randomly initialized learnable matrices~\cite{shao2022decoupled,wu2019graph,wu2020connecting,ye2021coupled,han2021dynamic,sun2022spatial,lu2020spatiotemporal,zhang2022adapgl}.
Two prominent models in this category are Graph WaveNet~\cite{wu2019graph} and MTGNN~\cite{wu2020connecting}, which have been widely applied or improved upon in subsequent works. In Graph WaveNet, the adaptive graph is produced as follows:
\begin{equation}
\tilde{\boldsymbol{A}}_{a d p}=\operatorname{SoftMax}\left(\operatorname{ReLU}\left(\boldsymbol{E}_1 \boldsymbol{E}_2^T\right)\right),
\end{equation}
where $\boldsymbol{E}_1\in \mathbb{R}^{N\times C}$ and $\boldsymbol{E}_2\in \mathbb{R}^{N\times C}$ are source node embedding and target node embedding, respectively. They are two learnable matrices with the random initialization, where $N$ denotes the number of nodes in the graph and $C$ denotes the dimension of the embedding. 

In contrast, the generation process of the adaptive graph in MTGNN is defined as:
\begin{gather}
% \begin{aligned}
\boldsymbol{M}_1  =\tanh \left(\alpha \boldsymbol{E}_1 \boldsymbol{\Theta}_1\right), \quad \boldsymbol{M}_2  =\tanh \left(\alpha \boldsymbol{E}_2 \boldsymbol{\Theta}_2\right), \\
\tilde{\boldsymbol{A}}_{a d p}  =\operatorname{ReLU}\left(\tanh \left(\alpha\left(\boldsymbol{M}_1 \boldsymbol{M}_2^T-\boldsymbol{M}_2 \boldsymbol{M}_1^T\right)\right)\right),
% \end{aligned}
\label{eq:adp}
\end{gather}
where $\boldsymbol{E}_1\in \mathbb{R}^{N\times C}$ and $\boldsymbol{E}_2\in \mathbb{R}^{N\times C}$ represent two randomly initialized node embeddings;  $\boldsymbol{\theta}_1$ and $\boldsymbol{\theta}_2$ are learnable parameters within the model; $\alpha$ is a hyperparameter for controlling the saturation rate of the activation function. Numerous subsequent random initialization-based adaptive graph learning methods are proposed based on the two aforementioned methods. For example, CCRNN~\cite{ye2021coupled} introduced a layer-wise adaptive graph learning mechanism to adjust the graph structures layer by layer. DMSTGCN~\cite{han2021dynamic} presented an adaptive graph learning approach with tensor decomposition. 

\textbf{Feature initialization-based} approaches aim to construct adaptive graph structure learning based on the given inputs or the hidden states~\cite{li2021dynamic,lan2022dstagnn,jin2022adaptive,peng2021dynamic,guo2020dynamic,shang2021discrete}. These models usually adopt learnable matrices or attention mechanism to incorporate with the given features for generating the adaptive graph structures. For example, DGCRN~\cite{li2021dynamic} proposed a recurrent adaptive graph learning mechanism based on the hidden states to construct the graph structures for each time step. 
% DSTAGNN~\cite{lan2022dstagnn} presented a self-attention-based  adaptive graph learning method to establish the connection between the graph structures and the hidden states. 
% BSTGCN~\cite{fu2020bayesian} designed a Bayesian graph learning mechanism based on the predefined graphs and input features. 
GTS~\cite{shang2021discrete} presented a novel probabilistic graph structure learning method based on input features.   

\subsubsection{Muti-Scale Spatial Learning}
Due to the wide existence of spatial heterogeneity in urban systems, entities can be divided into communities with different functions. Entities in the same community may have inter-community correlations, while entities in different communities could also have cross-community correlations. In light of these facts, some recent methods have investigated multi-scale spatial learning based on community partitioning. These methods often leverage domain knowledge to guide the community partitioning process.

In this research line, some studies obtain partitioned communities by artificial division~\cite{zhou2020foresee,zhou2020riskoracle} or clustering algorithms~\cite{guo2021hierarchical,wang2022multivariate,jin2022deep} while others obtain them by neural networks~\cite{chen2021group,tang2022hgarn,xia2021spatial}. 
For example, ST-SHN~\cite{xia2021spatial} and ST-HSL~\cite{li2022spatial} learn the hyperedges, \emph{i.e.} communities, of the hypergraph to capture global spatial dependencies for crime prediction. Besides, GAGNN~\cite{chen2021group} is a group-aware STGNN model for air quality prediction among hundreds of Chinese cities.
% , as shown in Figure~\ref{fig:multi_spatial_scale}. 
This model first proposed a differentiable grouping network for learning the assignment matrix, which automatically computes the mapping relationships between cities and city groups. 
% Then the spatial GCNs are respectively calculated for graph data at these two different scales to learn the inter-community and cross-community spatio-temporal dependencies. 
Another notable research line involves THINK~\cite{agarwal2022think} and DMGCRN~\cite{qin2021dmgcrn}, which utilize hyperbolic graph neural networks on the Poincare ball to capture multi-scale spatial dependencies more directly. The hyperbolic space is particularly suitable for modeling hierarchies, including local and global dependencies of spatio-temporal data, which makes it a promising approach for improving STGNN models.
% \begin{figure}[!t] 
% \centering
% % \vspace{-3mm}
% \includegraphics[width=0.48 \textwidth]{f13.png}
% \caption{The overview of GAGNN~\cite{chen2021group}.}
% \label{fig:multi_spatial_scale} % FIG
% % \vspace{-3mm}
% \end{figure}

\subsubsection{Heterogeneous spatial learning}
As mentioned in the introduction section, heterogeneity is an essential property of spatio-temporal data in smart cities wherein it displays varying patterns across various temporal or spatial ranges. Different from the above multi-scale spatial learning methods, some works focused on the fine-grained node-to-node heterogeneous relationships in the spatio-temporal data. 
To distinguish between the influence of static undirected edges (\emph{e.g.}, distance-based edges) and the dynamic directed edges (\emph{e.g.}, vehicle’s mobility-caused edges) in the spatio-temporal graph, HMGCN~\cite{wang2021forecasting} performs heterogeneous aggregation on the spatial dimension. Similarly, MasterGNN~\cite{han2021joint} constructs a heterogeneous graph structure based on multiple relations between air quality and weather monitoring stations, while HTGNN aggregates heterogeneous information from spatial-based intra-edges, temporal-based inter-edges, and spatio-temporal-based across-time-edges. 
Another line of heterogeneous spatial learning utilizes transportation, time, and geographical information to capture intricate spatio-temporal message passing. For instance, HeGA~\cite{liu2022hega} and MOHER~\cite{zhou2021modeling} design multiple transportation mode-based heterogeneous graphs to receive information from multi-sources at the same time, \emph{e.g.}, bike, bus, vehicle, etc. 
% The framework of MOHER is depicted in Figure~\ref{fig:heterogeneous_spatial}, where the spatio-temporal heterogeneous graph is constructed by the region pair-wise relations and inter-mode multi-relations to characterize the correlations between different transportation modes. Then the heterogeneous graph convolution operator is incorporated with LSTM to capture the complex spatio-temporal dependencies.
% Additionally, DH-GEM~\cite{guo2022talent} proposed the incorporation of node-position edges, while CAP~\cite{chen2019cap} further expanded this concept by designing node-time and node-location edges within the heterogeneous graph, allowing for the derivation of time and geographical knowledge.
% \begin{figure}[!b] 
% \centering
% % \vspace{-3mm}
% \includegraphics[width=0.48 \textwidth]{f24.png}
% \caption{The overview of MOHER~\cite{zhou2021modeling}.}
% \label{fig:heterogeneous_spatial} % FIG
% % \vspace{-3mm}
% \end{figure}

\subsection{Temporal Learning Methods}

\subsubsection{Multi-Scale Temporal Learning}
Given the prevalence of short- and long-range correlations in spatio-temporal data, capturing multi-scale temporal correlations has emerged as a crucial direction for improving temporal learning. So far, there are two mainstream design directions for multi-scale temporal learning in STGNNs. 
The first direction utilizes TCNs with receptive fields of varying scales~\cite{wu2020connecting,rathore2021multi}. A typical example is MTGNN~\cite{wu2020connecting} which employs multiple TCNs with various kernel sizes for learning temporal dependencies in different scales.
The second direction involves integrating other temporal learning networks~\cite{wang2020traffic,jin2022deep,li2021spatial}. 
For example, DMVST-VGNN~\cite{jin2022deep} jointly utilizes TCNs and Transformers for long-short range temporal learning.  
% As depicted in Figure~\ref{fig:multi_temporal_scale}, Traffic STGNN~\cite{wang2020traffic} achieves multi-scale temporal learning by multi-networks integration. This model adopts GRU for short-range temporal learning and Transformer for long-range temporal learning.   
% \begin{figure}[!t] 
% \centering
% % \vspace{-3mm}
% \includegraphics[width=0.47 \textwidth]{f14.png}
% \caption{The overview of Traffic STGNN~\cite{wang2020traffic}.}
% \label{fig:multi_temporal_scale} % FIG
% % \vspace{-3mm}
% \end{figure}

\subsubsection{Multi-Granularity Temporal Learning}
\begin{figure}[h] 
\centering
% \vspace{-3mm}
\includegraphics[width=0.35 \textwidth]{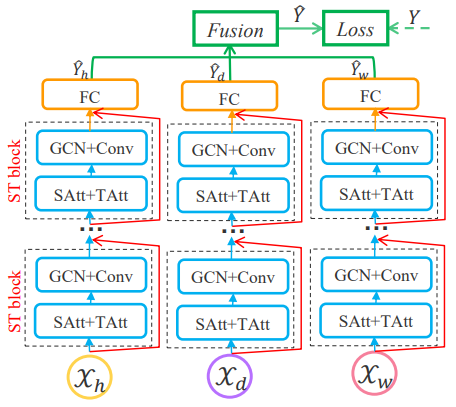}
\caption{The overview of ASTGCN~\cite{guo2019attention}.}
\label{fig:multi_time} % FIG
% \vspace{-1mm}
\end{figure}

There are multiple types of temporal characteristics in spatio-temporal data. For instance, the traffic flow at a given time is not only related to the recent traffic flow  but may also exhibit similarities to the traffic flow at the same time on the previous day or even the previous week. This reflects the closeness, periodicity, and trend, respectively. 
To consider the temporal characteristics at these three granularities, many previous works~\cite{guo2019attention,sun2020predicting,zhang2021traffic,zhang2020spatial,wang2022hierarchical} adopted a three-branch architecture to learn features from different temporal granularities separately, and then fuses the learned hidden states for predictions. As shown in Figure~\ref{fig:multi_time}, ASTGCN~\cite{guo2019attention} employs a typical three-branch architecture for multi-granularity temporal learning, where $\mathcal{X}_h$, $\mathcal{X}_d$ and $\mathcal{X}_w$represent the spatio-temporal data for the latest one hour, the data of the same hour from the previous day, and the data of the same hour from the previous week, respectively.  After going through separate branches, they are finally fused by the learnable weight matrix. 

\subsubsection{Decomposition Temporal Learning}
Individual temporal patterns usually contain a variety of hidden components, such as inherent components, diffusion components, and periodic components. To better capture these complex temporal dependencies, decomposition-based temporal learning methods have been proposed, which can automatically decompose and integrate different temporal components through special neural designs~\cite{oreshkin2021fc,fang2021spatio,shao2022decoupled,cao2020spectral,zheng2021hierst}. 
%Among them, HierST~\cite{zheng2021hierst}, FC-GAGA~\cite{oreshkin2021fc} and StemGNN~\cite{cao2020spectral} leveraged the N-BEATS~\cite{oreshkin2019n} model to decompose the different components from the individual time series. 
FC-GAGA~\cite{oreshkin2021fc} is a noteworthy example of decomposition methods that adopt the subtraction residual from N-BEATS~\cite{oreshkin2019n} to decompose different components in traffic data and model spatial correlations of each component. 
% As shown in Figure~\ref{fig:dtl}, FC-GAGA consists of multiple stacked layers, each consisting of a time gate block, a graph gate block, and several fully connected blocks.The time gate block aims to eliminate node-specific multiplicative seasonality from the block's input and reuse it at the block's output, while the graph gate block aims to capture spatial correlations from various entities. The fully connected blocks are similar to those in N-BEATS, acting on the final model output and removing redundant temporal components for downstream blocks through the two branches of forecast projection and backcast projection.
% \begin{figure}[!b] 
% \centering
% % \vspace{-3mm}
% \includegraphics[width=0.47 \textwidth]{f22.png}
% \caption{The overview of FC-GAGA~\cite{oreshkin2021fc}.}
% \label{fig:dtl} % FIG
% % \vspace{-3mm}
% \end{figure}
In addition to FC-GAGA, other works also adopted decomposition-based ideas. For example, StemGNN~\cite{cao2020spectral} decomposed the temporal components by the subtraction residual from N-BEATS, but modeled spatial correlations in the spectral domain. D2STGNN~\cite{shao2022decoupled} proposed a temporal residual decomposition method to incorporate with graph structure learning. STWave~\cite{fang2021spatio} directly utilized the discrete wavelet transform to disentangle the event and the trend from the spatio-temporal graph data.

\subsection{Spatio-Temporal Fusion Methods}

%\subsubsection{Spatio-Temporal Synchronous Graph Modeling}
\subsubsection{Spatio-Temporal Joint Modeling}
In Section~\ref{sec:stfa}, we have discussed basic spatio-temporal fusion architectures of STGNNs, which are either factorized or coupled by spatial learning networks and temporal learning networks.
Although these architectures can effectively learn spatial and temporal dependencies separately, they lack the ability to model the joint spatial-temporal dependencies, making it challenging to capture complex spatio-temporal relations across different time steps.

In recent years, some literature focused on jointly modeling spatial-temporal dependencies based on 3D GCN~\cite{xia20213dgcn}, Spatio-Temporal Joint GCN (STJGCN)~\cite{zheng2021spatio} and Spatio-Temporal Synchronous GCN (STSGCN)~\cite{song2020spatial}.  
Among them, STSGCN has become a mainstream method for spatio-temporal dependencies jointly fusion.
This type of neural architecture enables the modeling of spatio-temporal dependencies in a unified graph structure, which can replace the separated spatial learning networks and temporal learning networks. The crucial part of STSGNN is the construction of the spatio-temporal synchronous graph.
% , as shown in Figure~\ref{fig:sts}. 
The original spatio-temporal synchronous graph is simple, whose nodes with the same location are connected to each other across adjacent time steps. This graph construction approach not only characterizes spatial neighbors, but also temporal neighbors, establishing unified spatio-temporal relations. After graph construction, STSGNN employs a simple GCN model to capture the spatio-temporal dependencies.
% \begin{figure}[h] 
% \centering
% \vspace{-3mm}
% \includegraphics[width=0.45 \textwidth]{f16.png}
% \caption{Construction of the spatio-temporal synchronous graph~\cite{song2020spatial}. $A^{(t_i)}$ denotes the adjacency matrix of the spatial graph at time step $i$. $A^{t_i\rightarrow t_j}$ denotes the connections between the nodes with themselves at the time step $i$ and $j$.}
% \label{fig:sts} % FIG
% \vspace{-3mm}
% \end{figure}

There are some follow-up works~\cite{wu2022traversenet,li2021spatial,wang2022synchronous,li2022automated,jin2022automated,li2021multi,fang2021cdgnet,fang2022learning} based on STSGCN to further improve the spatio-temporal synchronous graph modeling in recent years. For example, STFGNN~\cite{li2021spatial} proposes to construct the spatio-temporal synchronous graph using not only topology-based but also similarity-based graphs, thereby making the spatio-temporal synchronous graph more informative. On the other hand, S2TAT~\cite{wang2022synchronous} proposes a spatio-temporal synchronous Transformer framework that employs attention mechanisms to enhance the learning capability.

\subsubsection{Automated Spatio-Temporal Fusion}
Given the complexity of STGNNs, designing optimal neural architectures can be a challenging task. Existing spatio-temporal fusion methods are usually designed empirically, and may not generalize well to different data scenarios due to the various spatio-temporal attributes present in different scenarios. Neural architecture search (NAS) methods offer opportunities for automated spatio-temporal fusion in STGNNs, and have shown promising results in discovering optimal architectures for various applications. 
%We can treat different spatial learning networks or temporal learning networks in STGNN as different blocks, and how these blocks are combined can be learned by the NAS methods. 

% Inspired by AutoST~\cite{li2020autost} for grid-based traffic prediction, 
AutoSTG~\cite{pan2021autostg} presented the first attempt to involve DARTS~\cite{liu2018darts}  (\emph{i.e.}, the most classical gradient-based NAS method) into STGNN. In AutoSTG, the whole neural network is divided into different stacked cells, and these cells are the basic units to perform NAS.
% , as shown in Figure~\ref{fig:auto}. 
% In the search phase, DARTS obtains the representation of each intermediate hidden state through a probabilistic parameterization method as follows:
% \begin{equation}
% \mathcal{H}^j=\sum_{i<j} \sum_{o \in \mathcal{O}} \frac{\exp \left(\alpha_o^{(i, j)}\right)}{\sum_{o^{\prime} \in \mathcal{O}} \exp \left(\alpha_{o^{\prime}}^{(i, j)}\right)} o\left(\mathcal{H}^i\right),
% \end{equation}
% where $\mathcal{H}^i$ denotes the $i$-th intermediate hidden state; $\mathcal{O}$ indicates the operation set; $o$ is the specific operation in $\mathcal{O}$; $\alpha_o^{(i, j)}$ denotes the architecture parameter from $i_{th}$ to $j_{th}$ hidden state. When the search phase finishes, the neural architecture is fixed according to the operation with the highest likelihood $\alpha_o^{(i, j)}$.
% \begin{figure}[h] 
% \centering
% \vspace{-3mm}
% \includegraphics[width=0.45 \textwidth]{f17.png}
% \caption{The overview of neural architecture search space in AutoSTG~\cite{pan2021autostg}. Zero, identity, spatial graph convolution (SC) and temporal convolution (TC) are four candidate operations in the search space.}
% \label{fig:auto} % FIG
% \vspace{-1mm}
% \end{figure}
Following AutoSTG, a series of studies~\cite{jin2023dual,jin2022automated,wang2022auto,li2022automated,wu2021autocts,jin2021hierarchical,jin2023urban,ke2023autostg+} to integrate NAS into STGNNs in recent years. For example, AutoSTS~\cite{li2022automated} integrates NAS into the spatio-temporal synchronous graph neural networks for searching the optimal architecture of different GCNs and TCNs. Likewise, Auto-DSTSGN~\cite{jin2022automated} also integrates NAS into the spatio-temporal synchronous graph neural networks, but focuses on searching the optimal adjacency matrices of spatio-temporal synchronous graphs. 
\section{Advanced Learning Frameworks}\label{sec:advanced}
In recent years, more and more advanced learning frameworks have been developed to enhance the performance of STGNN in terms of deep representation and prediction accuracy. In this section, we review and discuss some typical advanced learning frameworks that are combined with STGNNs.%, including adversarial learning, meta learning, self-supervised learning, continuous spatio-temporal modeling, physics-informed learning, and transfer learning.

\subsection{Adversarial Learning}
As traditional loss functions, such as L1 and L2 norms, are commonly used to measure prediction errors, they may lack the ability to capture the distribution and correlation between the predictions and real data. This limitation could potentially lead to distorted prediction results.
Hence, the adversarial loss can be introduced to incorporate with the traditional loss for addressing this problem to some extent, which has been widely applied in time series prediction. To incorporate the adversarial loss, Generative Adversarial Networks (GANs) have been proposed, with the neural predictors as the generators and the neural architecture of discriminators designed separately. We have witnessed a blossom of works~\cite{khaled2022tfgan,huang2022gan,jin2022gan,han2021joint,liu2022foreseeing,miao2022mba} to combine the adversarial loss with STGNNs for predictive learning tasks. 

\begin{figure}[h] 
\centering
\vspace{-2mm}
\includegraphics[width=0.45 \textwidth]{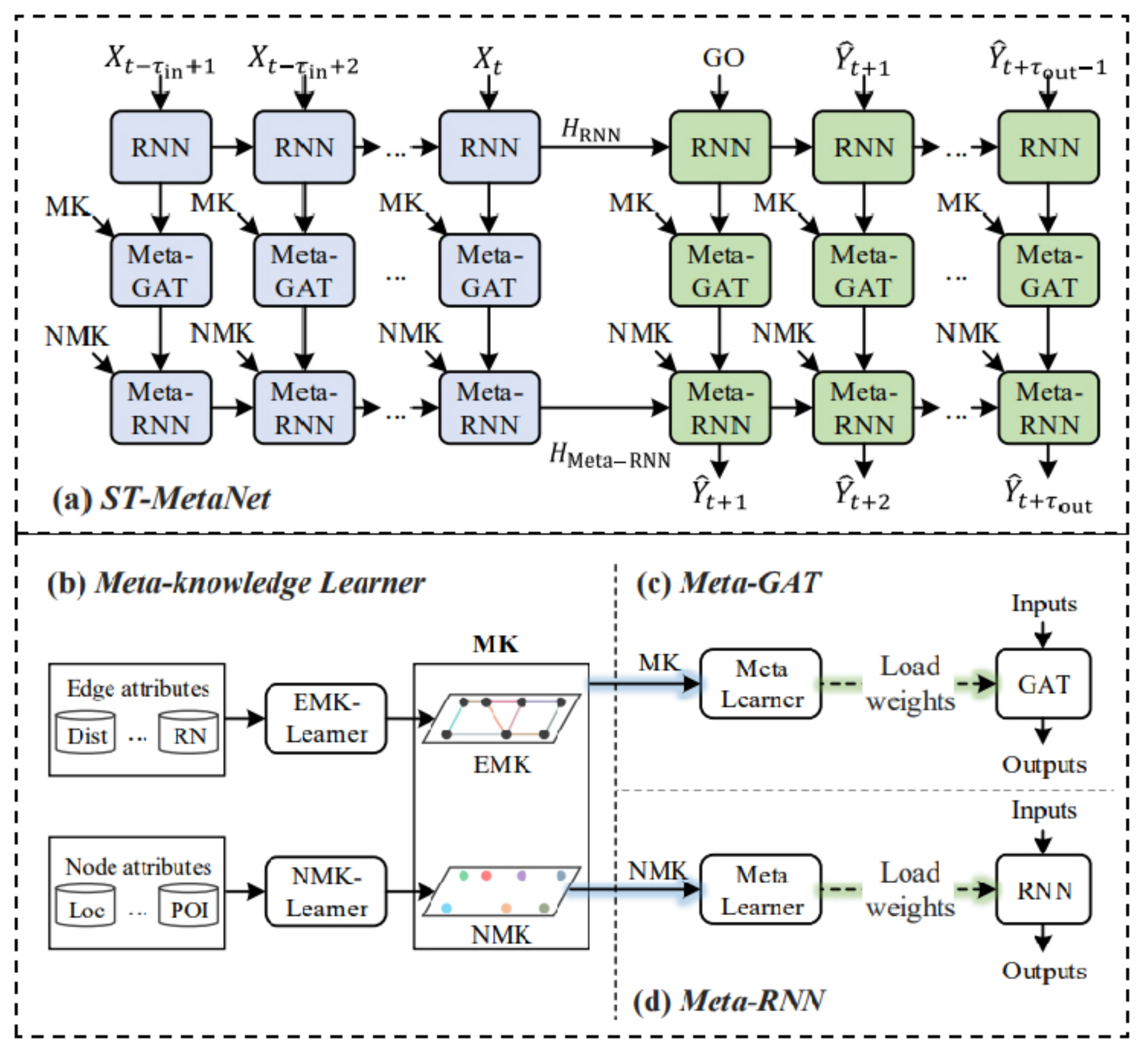}
\caption{The overview of ST-MetaNet~\cite{pan2019urban}.}
\label{fig:meta} % FIG
\vspace{-2mm}
\end{figure}
\subsection{Meta Learning}
Meta learning is an advanced learning paradigm focusing on the concept of ``learning to learn''. Incorporating meta learning techniques in STGNN models is important since they can capture high-dimensional heterogeneity and dynamic spatio-temporal dependencies from raw data, and teaching them how to learn can significantly improve their prediction performance. Typically, meta-learning-based STGNNs involve extracting additional spatio-temporal attributes through a meta-learner.
ST-MetaNet~\cite{pan2019urban} (see Figure~\ref{fig:meta}) is the pioneering study to introduce meta learning into STGNNs, which is composed of RNN, Meta-GAT, and Meta-RNN, and utilizes two types of meta-knowledge learners, namely Node Meta-Knowledge (NMK) and Edge Meta-Knowledge (EMK) learners, to effectively incorporate additional spatio-temporal information. 
% Both of the two different meta-knowledge learner use fully-connected network as the basic learning network. The NMK learner is designed to learn meta-knowledge from the node attributes, such as distances and road networks, while the EMK learner is designed to learn meta-knowledge from the edge attributes, such as locations and POIs. The learned meta-knowledge is then utilized to learn the weights of the Meta-RNN and Meta-GAT components, improving the overall learning capability of the model. 
% Formally, the calculation process of Meta-RNN for any spatial node $i$ is formulated as:
% \begin{equation}
% \begin{aligned}
% W_{\Omega}^{(i)} & =g_{W_{\Omega}}\left(\mathrm{NMK}\left(v^{(i)}\right)\right), \\
% U_{\Omega}^{(i)} & =g_{U_{\Omega}}\left(\mathrm{NMK}\left(v^{(i)}\right)\right), \\
% b_{\Omega}^{(i)} & =g_{b_{\Omega}}\left(\mathrm{NMK}\left(v^{(i)}\right)\right), \\
% h_t^{(i)} & =\operatorname{GRU}\left(z_t^{(i)}, h_{t-1}^{(i)} \mid W_{\Omega}^{(i)}, U_{\Omega}^{(i)}, b_{\Omega}^{(i)}\right),
% \end{aligned}
% \end{equation}
% where $W_{\Omega}^{(i)}, U_{\Omega}^{(i)}$ and $b_{\Omega}^{(i)}$ are learnable parameters in GRUs and they are generated from node attributes $v^{(i)}$ by node knowledge meta learner. The meta learner is composed of three different fully-connected networks $g_{W_{\Omega}}$, $g_{U_{\Omega}}$ and $g_{b_{\Omega}}$.  

In light of the success of ST-MetaNet, some other STGNN models have been proposed that incorporate meta learning. For example, ST-MetaNet+~\cite{pan2020spatio} fuse the dynamic spatio-temporal state and meta-knowledge for weight generation of GAT and GRU. AutoSTG~\cite{pan2021autostg} also adopts a meta learning method similar to ST-MetaNet while introducing neural architecture search, using meta-knowledge to generate weight parameters for graph convolution and temporal convolution. 
MegaCRN~\cite{jiang2022spatio} introduced an attention-based memory network, which stores the typical features in seen samples for further pattern matching, thus improving the capability of graph structure learning. In addition, meta learning can also be used for spatio-temporal graph knowledge transfer in predictive learning scenarios~\cite{lu2022spatio,mo2022cross}.

\subsection{Self-Supervised Learning}
Self-supervised learning is a type of method that transforms an unsupervised learning task into a supervised task by constructing its own labels.  The goal of this learning paradigm is to learn better representations for downstream supervised tasks. By using self-supervised learning, a representation with strong generalization performance can be learned. Combining STGNN models with self-supervised learning can enhance the capability of spatio-temporal graph learning, which can improve the accuracy of downstream predictive learning tasks.  

\begin{figure}[!t] 
\centering
\vspace{-3mm}
\includegraphics[width=0.47 \textwidth]{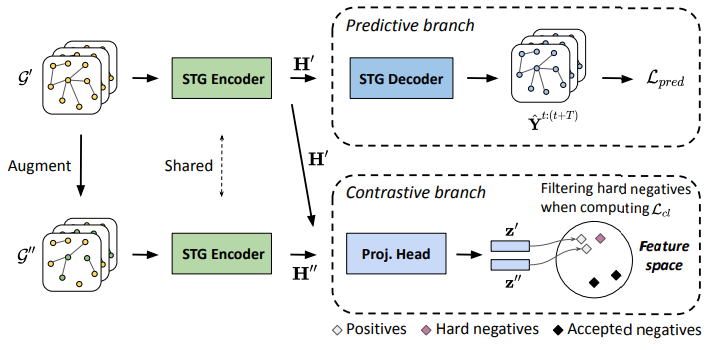}
\caption{The overview of STGCL~\cite{liu2022contrastive}.}
\label{fig:contrastive} % FIG
\vspace{-2mm}
\end{figure}

\textbf{Contrastive learning} is one of the most important self-supervised learning methods realized by constructing positive and negative samples, which has been introduced into STGNN models in recent years. One notable example is STGCL, which was introduced by Liu et al.~\cite{liu2022contrastive} and was the first work to incorporate contrastive learning into STGNN architectures.
As shown in Figure~\ref{fig:contrastive}, the first step of STGCL is the data augmentation to construct the positive and negative samples, where positive and negative samples are constructed using techniques such as edge masking, input masking, and temporal shifting.  After obtaining the positive and negative samples, the same STG encoder is employed to learn the spatio-temporal graph representation for both the original data and the augmented data.
Then, STGCL splits into two branches -- a predictive branch and a contrastive branch. In the predictive branch, the STG decoder directly outputs the prediction results and traditional data point errors, such as mean absolute error (MAE), can be used as the loss function.
In the contrastive branch, the two types of representation $H^{'}$ and $H^{''}$ are put forward into the projection head to further obtain the latent representation $z^{\prime}$ and $z^{\prime\prime}$. For the two latent representation, the contrastive loss proposed in GraphCL~\cite{you2020graph} was adopted in this case
% , which is as follows:
% \begin{equation}
% \mathcal{L}_{c l}=\frac{1}{M} \sum_{i=1}^M-\log \frac{\exp \left(\operatorname{sim}\left(\mathbf{z}_i^{\prime}, \mathbf{z}_i^{\prime \prime}\right) / \tau\right)}{\sum_{j \in \chi_i} \exp \left(\operatorname{sim}\left(\mathbf{z}_i^{\prime}, \mathbf{z}_j^{\prime \prime}\right) / \tau\right)},
% \end{equation}
% where $\operatorname{sim}(\cdot)$ denotes the cosine similarity, $\tau$ denotes  the temperature parameter. Note that, STGCL also proposed to filter out unsuitable negatives based on the unique properties of spatio-temporal graph data such as first-order neighbors of each node, closeness and periodicity temporal patterns, due to their similarity in the latent space. Thus, $chi_i$ denotes the set of acceptable negatives for the $i_{th}$ object after negative filting.

Based on STGCL, several other contrastive learning methods have been proposed to enhance the learning capabilities of STGNN in recent years. For example, SPGCL~\cite{li2022mining} proposed to learn the informative relations by maximizing the distinguishing margin between positive and negative neighbors for generating an optimal graph structure. ST-SSL~\cite{ji2022spatio} proposed an adaptive augmentation method over the spatio-temporal graph data at both attribute and structure levels. START~\cite{jiang2023self} presented a spatio-temporal graph-based contrastive learning method for trajectory representation learning. This model proposed multiple negative trajectories construction methods such as trajectory trimming and road segments mask, to aid the STGNN model in achieving better performance in travel time prediction tasks.

\subsection{Continuous Spatio-Temporal modeling}
Most existing STGNN-based approaches capture the spatial and temporal dependencies in a discrete way, leading to discontinuous latent state trajectories and higher prediction errors. To address this problem, some research has focused on continuous spatio-temporal modeling. Motivated by the success of Neural Ordinary Differential Equation (Neural-ODE)~\cite{chen2018neural}, a well-known approach for continuous system modeling, STGNNs combined with Neural-ODE can improve the capability of spatio-temporal graph representation learning in a continuous manner. STGODE~\cite{fang2021spatial} was the first attempt to introduce Neural-ODE into STGNNs, however, it only considers integrating Neural-ODE with GCN and neglects continuous modeling for temporal patterns. To achieve a joint continuous modeling for spatio-temporal dependencies, MTGODE~\cite{jin2022multivariate} introduced the integration of Neural-ODE with graph convolution operators and temporal convolution operators 
% as depicted in Figure~\ref{fig:ode}, 
to enable continuous spatio-temporal encoding.
In addition, 
% Social ODE~\cite{wen2022social} extended the ODE-based STGNN to the scenario of multi-agent trajectory prediction. 
MixRNN+~\cite{liang2022mixed} combined Neural-ODE and RNN for continuous recurrent hidden state modeling. STG-NCDE~\cite{choi2022graph} developed a STGNN combined with the neural controlled differential equation (Neural-CDE) for better continuous modeling, compared with Neural-ODE-based methods.

\begin{figure}[!b] 
\centering
\includegraphics[width=0.48 \textwidth]{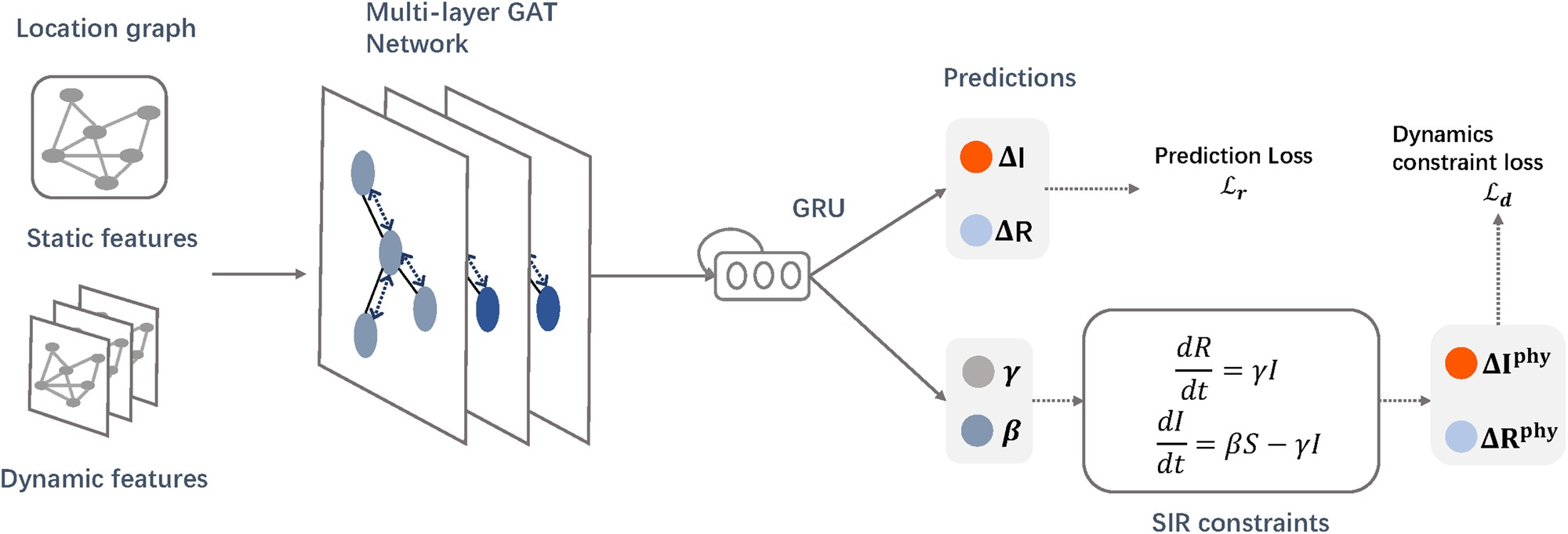}
\caption{The overview of STAN~\cite{gao2021stan}.}
\label{fig:physical} % FIG
\end{figure}
In addition to epidemic prediction tasks, there are also a few works in other domains. For example, STDEN~\cite{ji2022stden} proposed a unified framework that combines traffic potential energy field differential equations and neural networks for traffic flow prediction. 
\subsection{Physics-Informed Learning}
In the last few years, a new paradigm called Physics-Informed Neural Networks (PINNs)~\cite{karniadakis2021physics} have emerged for exploring and computing real-world dynamics integrating physical differential equations and neural networks with powerful fitting capabilities. The main advantage of PINNs is their ability to enforce physical constraints on the predictions, thereby ensuring that the model's outputs are consistent with the laws of physics.
Inspired by PINNs that are based on simple neural networks, physical-informed learning methods can be also combined with STGNNs, especially in epidemic prediction tasks~\cite{sun2022using,zheng2020spatial,la2020epidemiological,gao2021stan}. As shown in Figure~\ref{fig:physical}, STAN first integrates the constraints of SIR differential equations into the STGNN architecture. This model used GAT and GRU to capture the spatial and temporal dependencies respectively and performed a multi-task prediction. There are four components in the output of this model: transmission rate $\beta$, recovery rate $\gamma$, time-varying number of infections $\Delta \mathrm{I}$ and recoveries $\Delta \mathrm{R}$. These components need to satisfy physical constraints based on the SIR equation.
% , which can be expressed as follows:
% \begin{equation}
% \begin{aligned}   
% &\frac{d R}{d t}=\gamma I , \\
% &\frac{d I}{d t}=\beta S-\gamma I ,\\
% & S = N - I -R ,
% \end{aligned}
% \end{equation}
% where $S$ denotes the survivors and $N$ means the total number of people. In STAN, a constraint loss was used to enforce that the predicted time-varying infections and recoveries are close to those calculated by the SIR equations.

% The work~\cite{jia2021physics}  proposed a knowledge transfer approach from physics-based models to guide the learning of a recurrent graph convolution neural network for predicting flow and temperature in river networks.

\subsection{Transfer Learning}
Due to the scarcity of some spatio-temporal graph data, transfer learning techniques have become a cost-effective approach to extend the same basic STGNN model to different data scenarios. However, there are two main limitations in conducting transfer learning for STGNNs. The first one is the heterogeneity of spatial structures and the other one is the heterogeneity of temporal patterns in different circumstances. To be specific, in different scenarios, the spatial topology, relations, etc. are completely different as well as the temporal patterns such as periodicity and trend. 

The existing literature on spatio-temporal graph transfer learning can be roughly divided into three categories: clustering-based~\cite{mallick2021transfer,huang2021transfer,an2022hintnet}, domain adaptation-based~\cite{tang2022domain,liang2022cross} and meta-learning-based~\cite{lu2022spatio,mo2022cross}. 
For example, TL-DCRNN~\cite{mallick2021transfer} proposed a graph partitioning method to divide the entire highway network into different sub-clusters and then used the DCRNN model to learn the spatio-temporal dependencies from source sub-clusters to target sub-clusters. 
DASTNet~\cite{tang2022domain} combined the graph representation learning and multi-domains adversarial adaptation methods to obtain domain-invariant node embeddings, achieving the knowledge transfer among different scenarios with different spatial structures. 
% ST-GFSL~\cite{lu2022spatio} is a spatio-temporal graph neural network that employs Model-Agnostic Meta Learning (MAML) method for cross-city knowledge transfer. The first step in this model is to meta-train a base model on multiple source datasets, which generates the parameters for adaptation. During the adaptation phase, the feature extractor of the basic STGNN is initialized by the generated parameters, and then the parameters of the feature extractor and predictor are jointly trained on the target dataset.

% In order to facilitate future research in the field, we have compiled source codes of several representative STGNN models and categorized them according to the methods used to improve spatio-temporal representation learning. The models and corresponding categories are listed in Table~\ref{tab:code}. This resource can serve as a valuable reference for researchers seeking to develop and compare new STGNN models.

\section{Challenges and Future Directions}\label{sec:challenge}
We have investigated the applications, basic neural architectures, and recent advancements of STGNN for predictive learning in urban computing. Although STGNN models have achieved remarkable performance in recent years, there are still some challenging problems to be addressed, which point to potential future research directions. We summarize these challenges and suggest potentially feasible research directions as follows:
\begin{itemize}[leftmargin=*]
\item \textbf{Lack of interpretability:} So far, the vast majority of STGNN-related work has focused on improving predictive performance through sophisticated model design. However, research on the interpretability of models has been relatively lacking, that is, we cannot clearly understand which spatio-temporal features take a leading role in improving predictive performance. In the most recent work, STNSCM~\cite{deng2023spatio} proposed to construct a causal graph to describe the bike flow prediction and analyze the causal relationship between the spatio-temporal features and prediction results. 
%Causal-based spatio-temporal graph modeling could be a potential direction to enhance the interpretability of STGNN models.
In addition to the spatial perspective, some deep time series models have incorporated statistical modeling techniques to enhance the understanding of predictive outcomes~\cite{wa2019deep}. Hence, constructing interpretable STGNN models from both spatial and temporal perspectives is a potential direction.

\item \textbf{Lack of calibration methods:} How to establish trust among frontline urban managers regarding the predictive capabilities of STGNNs is a practical problem that needs further exploration. Hence, the significance of uncertainty quantification that can reflect the trustworthiness of prediction results needs to be emphasized. In order to improve the trustworthiness of the deep models, appropriate model calibration methods are necessary, which have been widely used in image recognition~\cite{ovadia2019can} and graph representation learning~\cite{wang2021confident} in recent years. At present, only works~\cite{wu2021quantifying,wen2023diffstg} have studied the uncertainty of STGNN models, and there is a lack of research on calibration methods. Calibration for the STGNN models need to take into account the characteristics of spatial and temporal simultaneously, thus it is more challenging than previous related works.

\item \textbf{Lack of physical constraints:} Most STGNN models capture the complex spatio-temporal dependencies through the integration of deep neural networks, while ignoring the consideration of physical constraints in different application domains, which makes the model less recognized in some professional fields. In recent years, although some STGNN models for epidemic prediction have combined professional differential equations as physical constraints~\cite{sun2022using,zheng2020spatial,la2020epidemiological,gao2021stan}, such work is still lacking and needs to be improved in other application fields.
%In recent years, differential equations with special physical meaning in some professional fields have been used in combination with neural networks to limit the predictions within a reasonable range, which are called as PINN.
\item \textbf{Lack of pre-training techniques:} Pre-training techniques have been greatly developed in the fields of time series and graph representation learning in recent years, but they are relatively lacking in STGNN-related work. In the most recent work, STEP~\cite{shao2022pre} proposed a pre-training model combined with the Mask Auto-Encoder (MAE)~\cite{he2022masked} architecture to efficiently learn temporal patterns from very long-term history spatio-temporal graph data. In the future, pre-training techniques for long-range spatial and long-term temporal learning are necessary, which are of great value to the scalability and deployability of the STGNN models.

\item \textbf{Lack of fine-grained NAS:} Due to the relatively complex components of STGNNs, automatically designing effective and reliable neural architectures is an urgent task. Although some existing works~\cite{jin2022automated,pan2021autostg,ke2023autostg+,wu2021autocts} have proposed the integration of NAS and STGNNs, most of them are all limited to coarse-scale and lack of search for fine-grained architectures in GNN (eg, aggregation methods, activation functions, etc.). Therefore, inspired by some current state-of-the-art NAS methods for GNNs~\cite{zhang2022deep,xu2023not}, proposing efficient fine-grained NAS for STGNNs is a promising direction.

\item \textbf{Hurdle of distribution shifts:} Spatio-temporal data%, such as traffic flows over road networks, 
are often collected from various locations and time periods, resulting in significant differences in the distribution of the training, validation, and test sets. 
% For instance, the training set may span the first two years, while the validation and test sets come from the following two years. 
For instance, we visualize the temporal distribution on Beijing Air Quality dataset. As shown in Figure~\ref{fig:shift}, the training data (periods A and B) and test data derive from different distributions, namely $P_{A}(x)\neq P_{B}(x)\neq P_{test}(x)$.
This can pose a challenge for STGNNs, as training a model on one dataset may not perform well on validation and test sets due to \emph{distribution shifts}, which is similar to the distribution shift issue in domain adaptation (where the joint distribution of inputs and outputs differs between the training and test stages). Despite its importance, this problem has received less attention in the spatio-temporal research community. While several studies~\cite{du2021adarnn} investigated defeating distribution shifts in time series, they fail to encode the spatial correlations among locations. 
\begin{figure}[!t] 
\centering
\includegraphics[width=0.35 \textwidth]{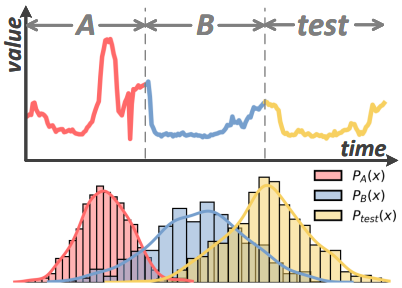}
\caption{Temporal Distribution Shift~\cite{xia2023deciphering}.}
\label{fig:shift} % FIG
\vspace{-4mm}
\end{figure}
\item \textbf{Exploring new training strategies}: Previous studies have primarily focused on introducing novel STGNNs with sophisticated layers or modules to enhance human mobility analytics. However, another promising direction is to investigate new training strategies. For instance, in traffic prediction tasks, every location is treated equally, and the data belonging to these locations are jointly fed into neural networks. Nevertheless, the complexity of modeling the spatio-temporal correlations of each location can vary significantly, necessitating a new training strategy such as curriculum learning. Curriculum learning trains a machine learning model on increasingly difficult data, starting from simpler data, and may be effective in addressing this issue. 
%In addition, other potential training strategies for STGNNs include multi-task learning, transfer learning, and continual learning. %By exploring new training strategies, we can improve the performance and accuracy of STGNNs and enable them to tackle even more complex tasks.

\item \textbf{Scalability issue}: One particularly challenging case for designing efficient STGNNs is when the number of locations in the sensor network is very large. For example, there are over ten thousand of loop detectors in PEMS systems. In this scenario, there is a need to develop STGNNs that can efficiently process and analyze the vast amounts of spatio-temporal data generated by the network while maintaining high prediction accuracy. Under this circumstance, more efficient AI solutions are appreciated, \emph{e.g.}, through model pruning/distillation, graph sampling techniques, or exploring the next-generation AI models with high efficiency. There are also a few studies probing into graph-free approaches \cite{liu2023we} to reduce computational costs when scaling up to large-scale sensor networks. 
In addition, some advancements in time series prediction research have also challenged the necessity of employing overly complex temporal learning models~\cite{zeng2023transformers,chen2023tsmixer}. Therefore, reducing the complexity of both spatial and timing computations to improve the scalability of STGNNs is a promising direction.
\end{itemize}

\vspace{-2mm}
\section{Conclusion}\label{sec:conclusion}
In this paper, we present a systematic survey of spatio-temporal graph neural networks (STGNNs) for predictive learning in urban computing. We start with a basic form and construction method of spatio-temporal graph data, and then summarize the predictive learning tasks involving STGNNs from different application domains in urban computing.
Next, Moving on, we delve into the fundamental neural network architectures that underpin STGNNs, including the spatial learning network and temporal learning network, which consist of graph neural networks (GNNs), recurrent neural networks (RNNs), temporal convolutional networks (TCNs), self-attention networks (SANs), and explore the basic fusion techniques used to integrate these spatio-temporal neural architectures.
To stay up-to-date with the latest developments in STGNNs, we review notable recent works, focusing on spatial learning methods, temporal learning methods, spatio-temporal fusion methods, and other advanced techniques that can be combined.
Finally, we summarize the challenges of current research and suggest some potential directions.

\section*{Acknowledgment}
This work is supported by Guangzhou Municiple Science and Technology Project 2023A03J0011. This project is also supported by the National Natural Science Foundation of China (62172034, 72242106), and the Beijing Natural Science Foundation (4212021). 

% %Dr. Reveryrand would like to acknowledge the funding by XLIM, Limoges, France. 
% The authors would like to thank Dr. David Root and Dr. Jean-Pierre Teyssier at Agilent Technologies for the loan of the time-domain nonlinear measurement equipment and TriQuint Semiconductor for the donation of the transistors. 

% if have a single appendix:
%\appendix[Proof of the Zonklar Equations]
% or
%\appendix  % for no appendix heading
% do not use \section anymore after \appendix, only \section*
% is possibly needed

% use appendices with more than one appendix
% then use \section to start each appendix
% you must declare a \section before using any
% \subsection or using \label (\appendices by itself
% starts a section numbered zero.)
%

% ============================================
%\appendices
%\section{Proof of the First Zonklar Equation}
%Appendix one text goes here %\cite{Roberg2010}.

% you can choose not to have a title for an appendix
% if you want by leaving the argument blank
%\section{}
%Appendix two text goes here.

% use section* for acknowledgement
%\section*{Acknowledgment}

%The authors would like to thank D. Root for the loan of the SWAP. The SWAP that can ONLY be usefull in Boulder...

% Can use something like this to put references on a page
% by themselves when using endfloat and the captionsoff option.
\ifCLASSOPTIONcaptionsoff
  \newpage
\fi

% trigger a \newpage just before the given reference
% number - used to balance the columns on the last page
% adjust value as needed - may need to be readjusted if
% the document is modified later
%\IEEEtriggeratref{8}
% The "triggered" command can be changed if desired:
%\IEEEtriggercmd{\enlargethispage{-5in}}

% ====== REFERENCE SECTION

%\begin{thebibliography}{1}

% IEEEabrv,

\bibliographystyle{IEEEtran}
\bibliography{IEEEabrv,Bibliography}
%\end{thebibliography}
% biography section
% 
% If you have an EPS/PDF photo (graphicx package needed) extra braces are
% needed around the contents of the optional argument to biography to prevent
% the LaTeX parser from getting confused when it sees the complicated
% \includegraphics command within an optional argument. (You could create
% your own custom macro containing the \includegraphics command to make things
% simpler here.)
%\begin{biography}[{\includegraphics[width=1in,height=1.25in,clip,keepaspectratio]{mshell}}]{Michael Shell}
% or if you just want to reserve a space for a photo:

% ==== SWITCH OFF the BIO for submission
% ==== SWITCH OFF the BIO for submission
\begin{IEEEbiography}[{\includegraphics[width=1.0in,height=1.5in,clip,keepaspectratio]{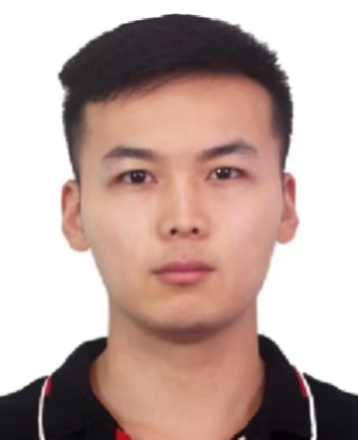}}]{Guangyin Jin} is an Assistant Research Fellow in National Innovative Institute of Defense Technology.
%, Academy of Military Sciences of PLA, Beijing, China. received a B.E. degree from the Department of Mechanical and Electrical Engineering of Xiamen University. 
He received a Ph.D degree at College of Systems Engineering of National University of Defense Technology. 
His research interest falls in the area of spatial-temporal data mining, graph neural networks and urban computing. So far, he has published more than ten papers in JCR Q1-level international journals such as TITS, TIST, TRC, INS, and top international conferences such as AAAI, CIKM, SIGSPATIL. 
%He also serves as the PC member or reviewer for top international conferences or journals such as AAAI, WWW, ECML-PKDD, TITS, TKDD, TRC, etc.
\end{IEEEbiography}

\begin{IEEEbiography}
[{\includegraphics[width=1in,height=1.25in,clip,keepaspectratio]{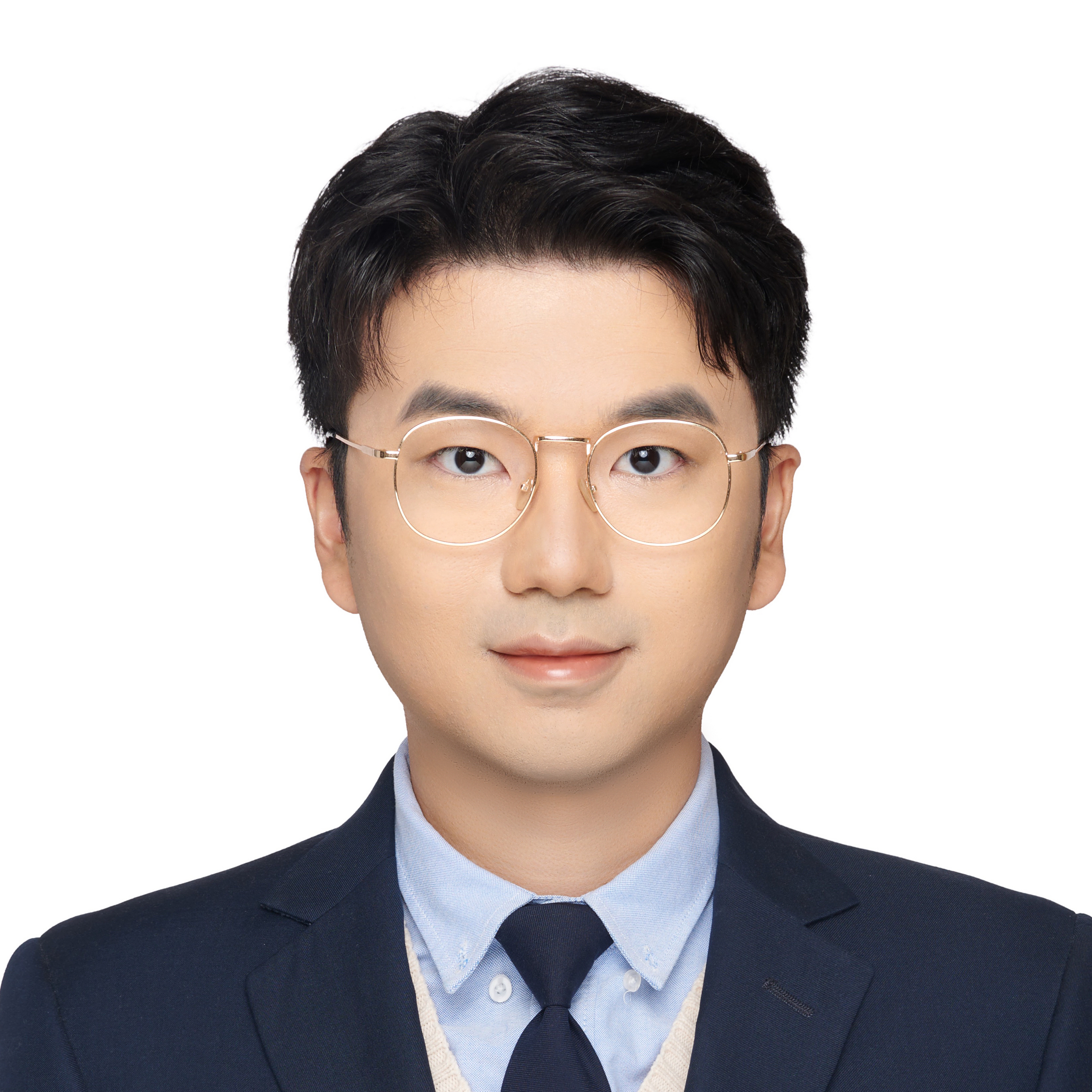}}]{Yuxuan Liang}
 is an Assistant Professor at Intelligent Transportation Thrust, Hong Kong University of Science and Technology (Guangzhou). He is currently working on the research, development, and innovation of spatio-temporal data mining and AI, with a broad range of applications in smart cities. Prior to that, he obtained his PhD degree at NUS. He published over 40 peer-reviewed papers in refereed journals and conferences, such as KDD, WWW, NeurIPS, ICLR, ECCV, AAAI, IJCAI, Ubicomp, and TKDE. Those papers have been cited over 2,000 times (Google Scholar H-Index: 21). He was recognized as 1 out of 10 most innovative and impactful PhD students focusing on data science in Singapore by Singapore Data Science Consortium (SDSC).
\end{IEEEbiography}

\begin{IEEEbiography}
[{\includegraphics[width=1.0in,height=1.5in,clip,keepaspectratio]{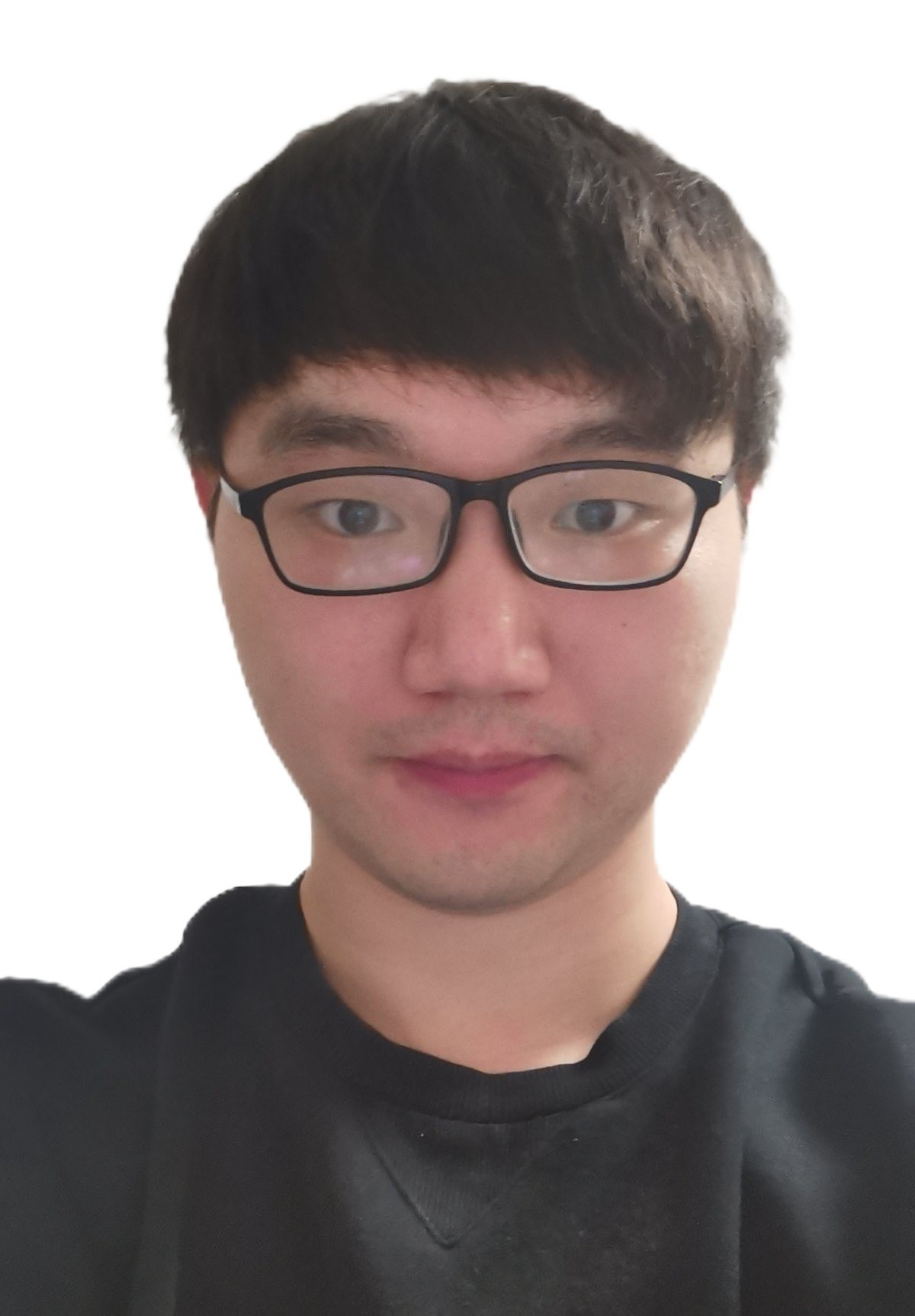}}]{Yuchen Fang} pursuing a Ph.D. at the University of Electronic Science and Technology of China. His general research interests are in spatio-temporal data mining, graph neural networks, and urban computing, with a special focus on traffic forecasting. He has published several papers in top journals and conference proceedings, such as ICDE, SIGIR, AAAI, and TITS.
\end{IEEEbiography}

\begin{IEEEbiography}
[{\includegraphics[width=1.0in,height=1.5in,clip,keepaspectratio]{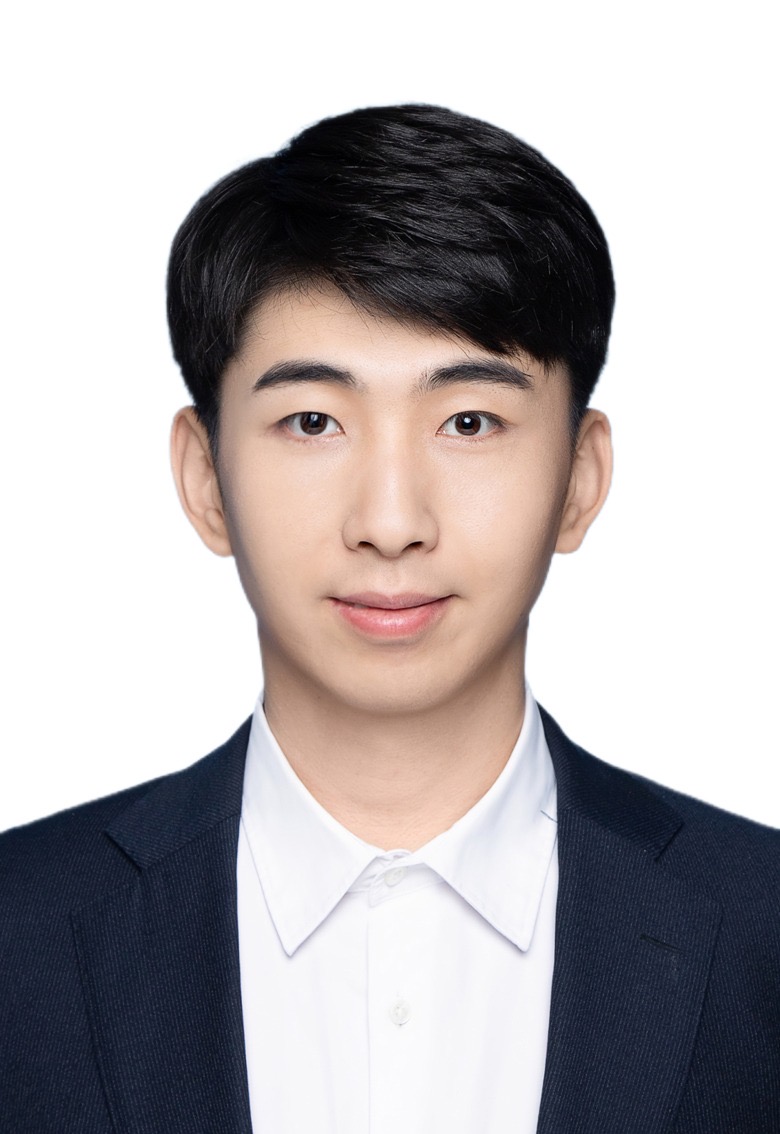}}]{Zezhi Shao} received the B.E. degree from Shandong University, Jinan, China, in 2019. He is currently pursuing a Ph.D. degree with the Institute of Computing Technology, Chinese Academy of Sciences, China. His research interests include traffic forecasting, multivariate time series forecasting, graph neural networks, and spatial-temporal data mining. He has published several papers in top journals and conferences, such as TKDE, SIGKDD, VLDB, CIKM, etc.
\end{IEEEbiography}

\begin{IEEEbiography}[{\includegraphics[width=1.0in,height=1.5in,clip,keepaspectratio]{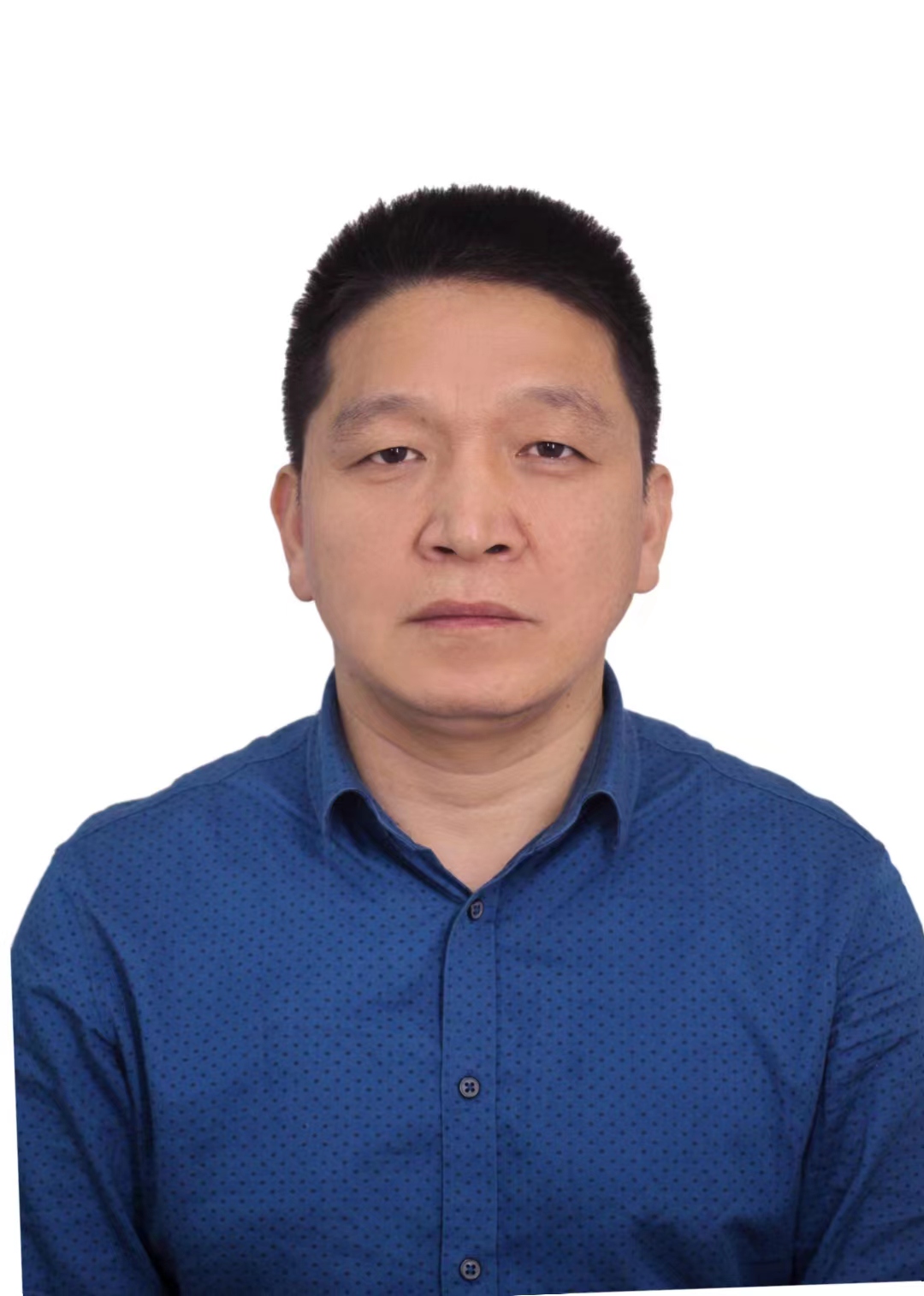}}]{Jincai Huang} is a professor of College of Systems Engineering of National University of Defense Technology, Changsha, Hunan, China. His main research interests include artificial general intelligence, deep reinforcement learning, and multi-agent systems. He has published more than 80 SCI / EI indexed papers, and he also serves as a director of the Machine Learning Committee of the Chinese Association for Artificial Intelligence (CAAI). 
\end{IEEEbiography}

\begin{IEEEbiography}
    [{\includegraphics[width=1in,height=1.25in,clip,keepaspectratio]{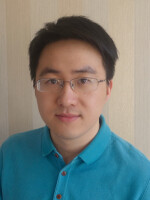}}] {Junbo Zhang} is a Senior Researcher of JD Intelligent Cities Research. He is leading the Urban AI Product Department of JD iCity at JD Technology, as well as AI Lab of JD Intelligent Cities Research. Prior to that, he worked at Micorsoft Research Asia from 2015 to 2018. He has published over 60 research papers in spatio-temporal data mining and AI, urban computing, deep learning, and federated learning. He serves as an Associate Editor of ACM Transactions on Intelligent Systems and Technology. He received a number of honors, including the Second Prize of the Natural Science Award of the Ministry of Education in 2021, the 22nd and 23rd China Patent Excellence Award in 2021 \& 2022, CCF Award for Science and Technology Award--Technological Progress Outstanding Award in 2021. He is an ACM/IEEE/CCF senior member. 
\end{IEEEbiography}

\begin{IEEEbiography}[{\includegraphics[width=1in,height=1.25in,clip,keepaspectratio]{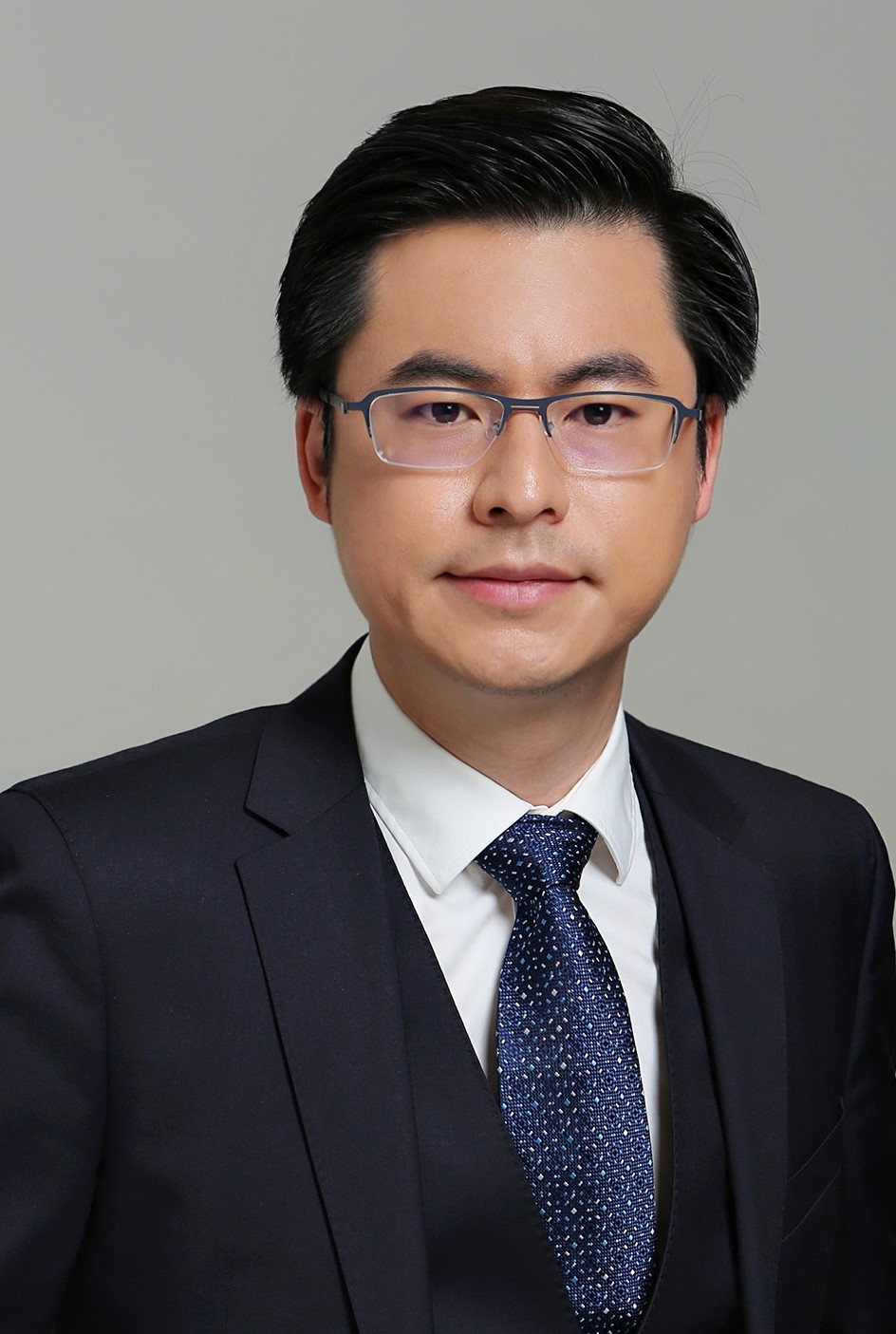}}]{Yu Zheng} (Fellow, IEEE) is the Vice President of JD.COM and head JD Intelligent Cities Research. Before Joining JD.COM, he was a senior research manager at Microsoft Research. He currently serves as the Editor-in-Chief of ACM Transactions on Intelligent Systems and Technology and has served as the program co-chair of ICDE 2014 (Industrial Track), CIKM 2017 (Industrial Track) and IJCAI 2019 (industrial track). He is also a keynote speaker of AAAI 2019, KDD 2019 Plenary Keynote Panel and IJCAI 2019 Industrial Days. His monograph, entitled Urban Computing, has been used as the first text book in this field. In 2013, he was named one of the Top Innovators under 35 by MIT Technology Review (TR35) and featured by Time Magazine for his research on urban computing. In 2016, Zheng was named an ACM Distinguished Scientist and elevated to an IEEE Fellow in 2020 for his contributions to spatio-temporal data mining and urban computing.
\end{IEEEbiography}

%% if you will not have a photo at all:
%\begin{IEEEbiographynophoto}{Ignacio Ramos}
%(S'12) received the B.S. degree in electrical engineering from the University of Illinois at Chicago in 2009, and is currently working toward the Ph.D. degree at the University of Colorado at Boulder. From 2009 to 2011, he was with the Power and Electronic Systems Department at Raytheon IDS, Sudbury, MA. His research interests include high-efficiency microwave power amplifiers, microwave DC/DC converters, radar systems, and wireless power transmission.
%\end{IEEEbiographynophoto}

%% insert where needed to balance the two columns on the last page with
%% biographies
%%\newpage

%\begin{IEEEbiographynophoto}{Jane Doe}
%Biography text here.
%\end{IEEEbiographynophoto}
% ==== SWITCH OFF the BIO for submission
% ==== SWITCH OFF the BIO for submission

% You can push biographies down or up by placing
% a \vfill before or after them. The appropriate
% use of \vfill depends on what kind of text is
% on the last page and whether or not the columns
% are being equalized.

\vfill

% Can be used to pull up biographies so that the bottom of the last one
% is flush with the other column.
%\enlargethispage{-5in}

% that's all folks
\end{document}